%% file: main.tex
\documentclass[10pt,twocolumn,letterpaper]{article}

\usepackage{cvpr}
\usepackage{times}
\usepackage{epsfig}
\usepackage{graphicx}
\usepackage{amsmath}
\usepackage{amssymb}
\usepackage{subfig}
\usepackage{authblk}

\usepackage[pagebackref=true,breaklinks=true,letterpaper=true,colorlinks,bookmarks=false]{hyperref}

\cvprfinalcopy % *** Uncomment this line for the final submission

\def\cvprPaperID{XXXX} % *** Enter the CVPR Paper ID here

\ifcvprfinal\fi
\begin{document}

% Ruizhe Wang, Chih-Fan Chen, Hao Peng, Xudong Liu, Oliver Liu, Xin Li
\author[1]{Ruizhe Wang \thanks{equal contribution} \thanks{\{ruizhe, chihfan, hpeng, xudong,oliver\}@oben.com}}
\newcommand\CoAuthorMark{\footnotemark[1]}
\newcommand\FirstEmailMark{\footnotemark[2]}
\author[1]{Chih-Fan Chen \protect\CoAuthorMark \protect\FirstEmailMark}
\author[1]{Hao Peng\protect\FirstEmailMark}
\author[1, 2]{Xudong Liu \protect\FirstEmailMark}
\author[1]{Oliver Liu  \protect\FirstEmailMark }
\author[2]{Xin Li \thanks{xin.li@mail.wvu.edu}}
\affil[1]{Oben, Inc}
\affil[2]{West Virginia University}

%%%%%%%%% TITLE

\title{Digital Twin: Acquiring High-Fidelity 3D Avatar from a Single Image}
\maketitle

\input{abstract.tex}

\input{introduction.tex}

\input{related_work.tex}

\input{proposed_method.tex}

\input{experimental_results.tex}

\input{conclusion.tex}

\clearpage
\pagebreak
\part*{Supplemental Material}
\input{supplemental_material.tex}

%\clearpage 
{\small
\bibliographystyle{ieee_fullname}
\bibliography{cvpr2020_twin}
}

\end{document}

%% file: abstract.tex
\begin{abstract}
    We present an approach to generate high fidelity 3D face avatar with a high-resolution UV texture map from a single image. To estimate the face geometry, we use a deep neural network to directly predict vertex coordinates of the 3D face model from the given image. The 3D face geometry is further refined by a non-rigid deformation process to more accurately capture facial landmarks before texture projection. A key novelty of our approach is to train the shape regression network on facial images synthetically generated using a high-quality rendering engine. Moreover, our shape estimator fully leverages the discriminative power of deep facial identity features learned from millions of facial images. We have conducted extensive experiments to demonstrate the superiority of our optimized 2D-to-3D rendering approach, especially its excellent generalization property on real-world selfie images. Our proposed system of rendering 3D avatars from 2D images has a wide range of applications from virtual/augmented reality (VR/AR) and telepsychiatry to human-computer interaction and social networks.
\end{abstract}

%% file: introduction.tex
\section{Introduction}
\label{sec:introduction}

    \begin{figure}[t]
      \includegraphics[width=\columnwidth]{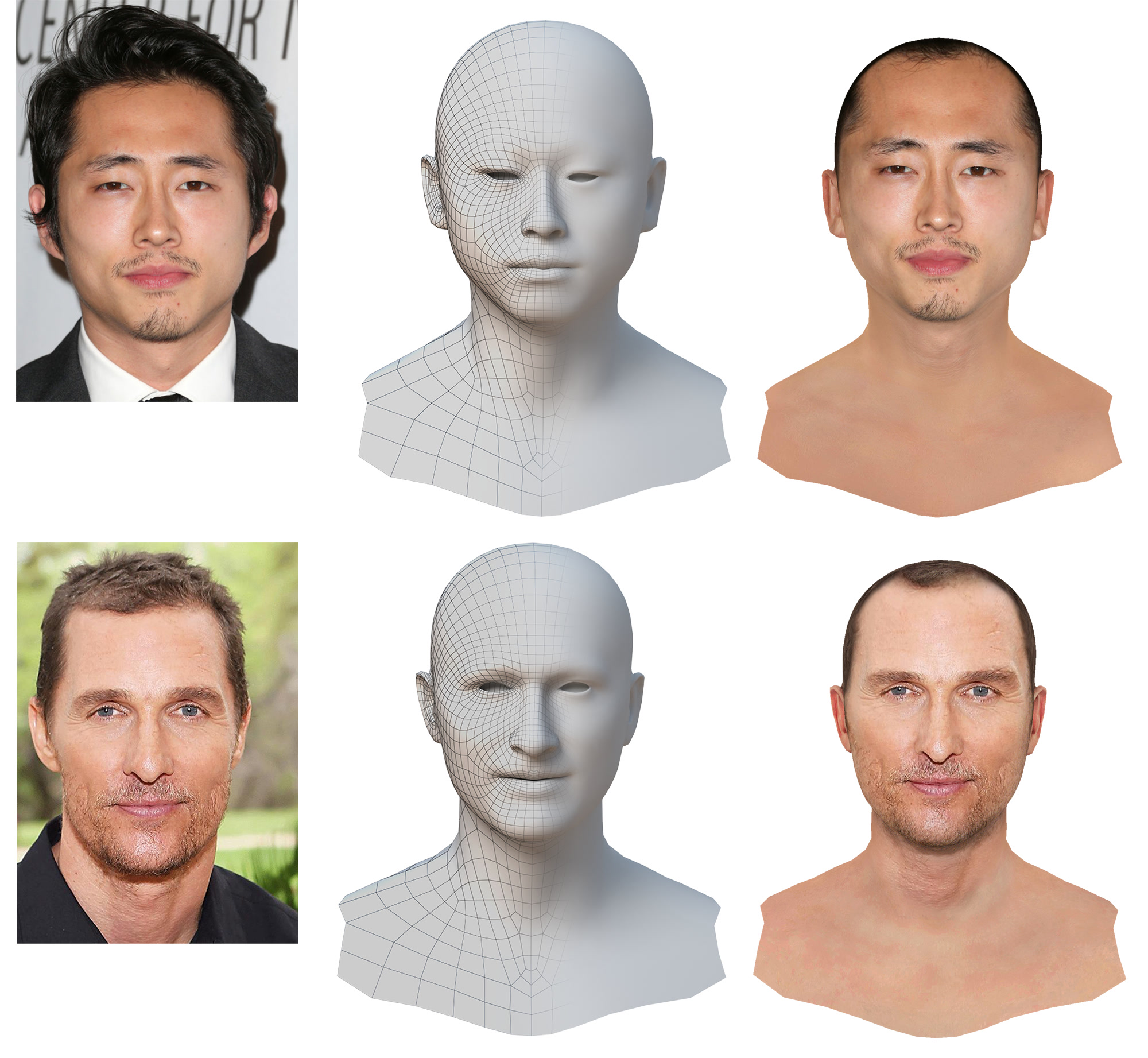}
      \caption{Sample outputs of our proposed avatar generation approach. From left to right: input image, inferred shape model with low polygon count, composite model with UV diffuse map}
      \label{fig:teaser}
    \end{figure}
    
    Acquiring high quality 3D avatars is an essential task in many vision applications including VR/AR, teleconferencing, virtual try-on, computer games, special effect, and so on. A common practice, adopted by most professional production studios, is to {\em manually} create avatars from 3D scans or photo references by skillful artists. This process is often time consuming and labor intensive because each model requires days of manual processing and touching up. It is desirable to automate the process of 3D avatar generation by leveraging rapid advances in computer vision/graphics and image/geometry processing. There has been a flurry of works on generating 3D avatars from handheld video \cite{ichim2015dynamic}, Kinect \cite{aitpayev2012creation} and mesh model \cite{wang2011text} in the open literature. 
    
    Developing a fully automatic system for generating 3D avatar from a single image is challenging because the estimation of both facial shape and texture map involves intrinsically ambiguous composition of light, shape and surface material. Conventional wisdom attempts to address this issue by {\em inverse rendering}, which formulates image decomposition as an optimization problem and estimates the parameters best fitting the observed image \cite{Blanz1999_3DMD, aldrian2012inverse,romdhani2005estimating}. More recently, several deep learning based approaches have been proposed - either in a supervised setup to directly regress the parameterized 3D face model \cite{zhu2016face,feng2018prn,tuan2018extreme} or in an unsupervised fashion \cite{Tewari2017MoFAMD,genova2018unsupervised,sanyal2019learning} with the help of a differentiable rendering process. However, these existing methods usually assume {\em over-simplified} lighting, shading and skin surface models, which does not take real-world complexities (\eg, sub-surface scattering, shadows caused by self-occlusion and complicated skin reflectance field \cite{debevec2000acquiring}) into account. Consequently, the recovered 3D avatar often does not faithfully reflect the actual face presented in the image.
    
    To meet those challenges, we propose a novel semi-supervised approach to  utilize synthetically-rendered, photo-realistic facial images augmented from a prioritized 3D facial scan dataset. Upon collecting and processing 482 neutral facial scans with a medical grade 3D facial scanner, we perform shape augmentation and utilize a high-fidelity rendering engine to create a large collection of photo-realistic facial images. To the best of our knowledge, this work is the first attempt to leverage photo-realistic facial image synthesis for accurate face shape inference.
    
    For facial geometry estimation, we propose to first extract deep facial identity features \cite{taigman2014deepface,parkhi2015deep}, {\em trained on millions of images}, which encodes each face into a unique latent representation, and regress the vertex coordinates of a generic 3D head model. To better capture facial landmarks for texture projection, the vertex coordinates are further refined in a non-rigid manner by jointly optimizing over camera intrinsic, head pose, facial expression and a per-vertex corrective field. Our final generated model consists of a shape model with low polygon counts but a high-resolution texture map with sharp details, which allows efficient rendering even on {\em mobile devices} (as shown in Fig.~\ref{fig:teaser}).
    
    At the system level, 3D avatars created by our approach are similar to those of {\em Pinscreen} \cite{hu2017avatar, yamaguchi2018high} and {\em FaceSoft.io} \cite{gecer2019ganfit}. For {\em Pinscreen} avatar, the shape model is reconstructed via an analysis-by-synthesis method \cite{Blanz1999_3DMD}; while our shape model is directly regressed from deep facial identity features, hence reaching {\em higher shape similarity}. For {\em FaceSoft.io} avatar, both shape and texture models utilize a collection of 10,000 facial scans \cite{Booth2016_3DMM10000}; while our semi-supervised method only uses 482 scans and is still capable of achieving similar shape reconstruction accuracy and {\em higher resolution} UV-texture map from an input selfie.
    
    Our key contributions can be summarized as follows:
    
    $\bullet$ A system for generating a high-fidelity UV-textured 3D avatar from a single image which can be efficiently rendered in real time even on mobile devices.
    
    $\bullet$ Training a shape estimator on the synthetic photo-realistic images by using pre-trained deep facial identity features. The trained networks demonstrate excellent generalization properties on real-world images.
    
    $\bullet$ Extensive qualitative and quantitative evaluation of the proposed method against other state-of-the-art face modeling techniques demonstrates its superiority (\ie, higher shape similarity and texture resolution).

%% file: related_work.tex
\section{Related Works}
\label{sec:related}

\textbf{3D Face Representation.} 3D Morphable Model (3DMM) \cite{Blanz1999_3DMD} uses Principal Component Analysis (PCA) on aligned 3D neutral faces to reduce the dimension of 3D face representation making the face fitting problem more tractable. The FaceWareHouse technique ~\cite{Cao2014_FWH} enhances the original PCA-based neutral face model with expressions by applying multi-linear analysis \cite{Vlasic2005_FTM} to a large collection of 4D facial scans captured with RGB-D sensors. The quality of multi-linear model was further improved in \cite{Bolkart2015_GML} by jointly optimizing the model and the group-wise registration of 3D scans. In \cite{Booth2016_3DMM10000}, a Large Scale Facial Model with 10,000 faces was generated to maximize the coverage of gender and ethnics. The training data was further enlarged in \cite{Li2017_4D}, which created a linear shape space trained from 4D scans of 3800 human heads. More recently, a non-linear model was proposed in \cite{Tran2018_nl3DMM} from a large set of unconstrained face images without the necessity of collecting 3D face scans.

\textbf{Fitting via Inverse Rendering.} Inverse rendering \cite{aldrian2012inverse, Blanz1999_3DMD} formulates 3D face modeling as an optimization problem over the entire parameter space seeking the best fitting for the observed image. In addition to pixel intensity values, other constraints such as facial landmarks and edge contours, are exploited for more accurate fitting \cite{romdhani2005estimating}. More recently, GanFit \cite{gecer2019ganfit} used a generative neural network for facial texture modeling and utilized an additional facial identity loss function in the optimization formulation. The inverse rendering based modeling approach has been widely used in many applications \cite{yamaguchi2018high,li2018avatar,thies2016face2face,geng20193d}.

\textbf{Supervised Shape Regression.} Convolutional Neural Network (CNN) based approaches have been proposed to directly map an input image to the parameters of a 3D face model such as 3DMM \cite{dou2017end,zhu2016face,jourabloo2016large,tuan2017regressing,yi2019mmface}. In \cite{jackson2017large}, a volumetric representation was learned  from an input image. In \cite{sela2017unrestricted}, an input color image was mapped to a depth image using an image translation network. In \cite{feng2018prn}, a network was proposed to jointly reconstruct the 3D facial structure and provide dense alignment in the UV space. The work of \cite{tuan2018extreme} took a layered approach toward decoupling low-frequency geometry from its mid-level details estimated by a shape-from-shading approach. It is worth mentioning that many CNN-based approaches use facial shape estimated by inverse rendering as the ground truth during training.

\textbf{Unsupervised Learning.} Most recently, face modeling from images via unsupervised learning becomes popular because it affords almost unlimited amount of data for training. An image formation layer was introduced in \cite{Tewari2017MoFAMD} as the decoder jointly working with an auto-encoder architecture for end-to-end unsupervised training. SfSNet \cite{sengupta2018sfsnet} explicitly decomposes an input image into albedo, normal and lighting components, which are then composed back to approximate the original input image. 3DMM parameters were first directly learned in ~\cite{genova2018unsupervised} from facial identity encoding and then the problem of parameter optimization was formulated in an unsupervised fashion by introducing a differentiable renderer and a facial identity loss on the rendered facial image. A multi-level face model, (\ie 3DMM with corrective field) was developed in~\cite{tewari2018self} following an inverse rendering setup that explicitly models geometry, reflectance and illumination per vertex. RingNet \cite{sanyal2019learning} employed a similar idea as the triplet loss for encoding all images of the same subject to the same latent shape representation. 

\textbf{Deep Facial Identity Feature.} Recent advances in face recognition \cite{taigman2014deepface, parkhi2015deep, schroff2015facenet} attempt to encode all facial images of the same subject under different conditions into identical feature representations, namely deep facial identity features. Several attempts have been made to utilize this robust feature representation for face modeling. GanFit \cite{gecer2019ganfit} used an additional deep facial identity loss to the commonly used landmark and pixel intensity losses. In \cite{genova2018unsupervised}, 3DMM parameters were directly learned from deep facial features. Although our shape regression network is similar to theirs, the choice of training data is different. Unlike their unsupervised setting, we opt to work with supervision by synthetically rendered facial images.

%% file: proposed_method.tex
\section{Proposed Method}
\label{sec:method}

\subsection{Overview}
\label{sec:overview}

\begin{figure*}[t]
          \includegraphics[width=\textwidth]{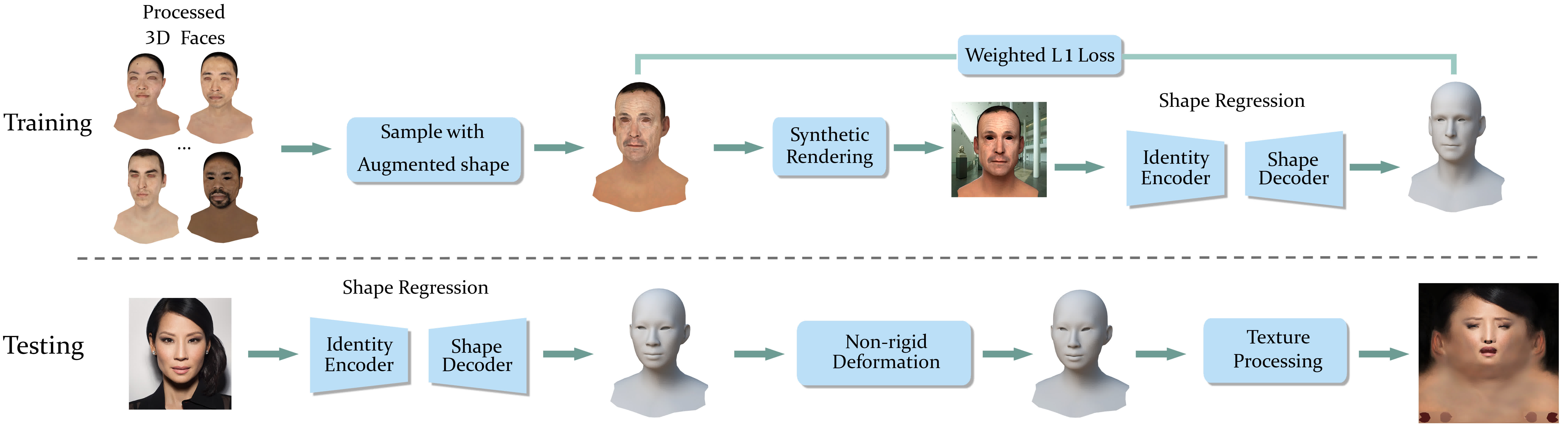}
          \caption{Overview of the proposed approach. During training, we learn a shape regression neural network on photo-realistic synthetic facial images. During testing, we infer a low polygon count shape model with a UV diffuse map generated from the projected texture.}
          \label{fig:overview}
    \end{figure*}

    An overview of the proposed method is shown in Fig.~\ref{fig:overview}. To facilitate facial image synthesis (Sec.~\ref{sec:photorealistic}) for training a shape regression neural network (Sec.~\ref{sec:regressing}), we have collected and processed a prioritized 3D face dataset, from which we can sample augmented 3D face shape with UV-texture to render a large collection of photo-realistic facial images. During testing, the input image is first used to directly regress the 3D vertex coordinates of a 3D face model with the given topology, which are furthered refined to fit the input image with a per-vertex non-rigid deformation approach (Sec.~\ref{sec:nonrigid}). Upon accurate fitting, selfie texture is projected to the UV space to infer a complete texture map (Sec.~\ref{sec:texture}).
    
\subsection{Photo-Realistic Facial Synthesis}
\label{sec:photorealistic}

    \textbf{3D Scan Database}. The most widely used Basel Face Model (BFM) \cite{Paysan2009_BFM} has two major drawbacks. First, it consists of 200 subjects but mainly Caucasian, which might lead to biased face shape estimation. Second, each face is represented by a dense model with high polygon count, per-vertex texture appearance and frontal face only, which limits its use for production-level real-time rendering. To overcome these limitations, we have collected a total of 512 subjects using a professional-grade multi-camera stereo scanner (3dMD LLC, Atlanta \footnote{\url{http://www.3dmd.com/}}) across different gender and ethnicity as shown in Table~\ref{tab:race}.
    
    A face representation containing a head model of 2925 vertices and a diffuse map sized by $2048 \times 2048$ is used. We take a non-rigid alignment approach \cite{Cao2014_FWH} of deforming a generic head model to match the captured facial scan. Then we transfer the texture onto the generic model's UV space. With further manual artistic touch up, we obtain the final high-fidelity diffuse map.

    \begin{table}
    \center
        \begin{tabular}{|c|c|c|c|c|}
        \hline
        \scriptsize{Gender/ Ethnicity}   & White     & Asian     & Black     & Total \\ \hline
        Male        & 82 / 5    & 178 / 5   & 8 / 5     & 268 / 15   \\ \hline
        Female      & 45 / 5    & 164 / 5   & 5 / 5     & 214 / 15   \\ \hline
        Total       & 127 / 10  & 342 / 10  & 13 / 10   & 482 / 30   \\ \hline
        \end{tabular}
        \caption{The distribution of gender and ethnicity in our database. Note that we randomly select 5 subjects for each group for testing and the rest subjects are used for training and validation.}
    \label{tab:race}
    \end{table}

    \textbf{Shape Augmentation}. 482 subjects are far from enough to cover all possible facial shape variations. While it is expensive to collect thousands of high-quality facial scans, we adopt an alternative shape augmentation approach to improve the generalization ability of the trained neural network. First, we adopt a recent deformation representation (DR) \cite{wu2018alive, gao2019sparse} to model a 3D facial mesh $\mathbf{P}$. DR feature encodes the $i$th vertex $\mathbf{P}^i=[P_x^i, P_y^i, P_z^i]$ as a $\mathbb{R}^9$ vector. Hence the DR feature of the entire mesh is represented as a vector $\mathbf{D}\in\mathbb{R}^{|\mathbf{P}|\times9}$. Please see the supplementary material on how to compute a DR feature $\mathbf{D}$ from $\mathbf{P}$ and vice versa.

    Upon obtaining a set of DR features as $(\mathbf{D}_1, ..., \mathbf{D}_N)$ where N is the total number of subjects, we follow \cite{jiang2019disentangled} to sample new DR features. More specifically, we sample a vector $(r, \theta_1, ..., \theta_{m_1})$ in Polar coordinates, where $r$ observes a uniform distribution $\mathbf{U}[0.6, 1.3]$ and $\theta_i$ follows uniform distribution $\mathbf{U}[0, \pi/2]$. We calculate its corresponding Cartesian coordinates $(a_1, a_2, ..., a_m)$ and interpolate the sampled DR features as $\sum_{i=1}^{m}a_i\mathbf{D}_i$, from which we further calculate the corresponding facial mesh. In our experiments, we use $m=5$ and only select samples from the same gender and ethnicity. We generate 10,000 new 3D faces with a ratio of $0.65/0.30/0.05$ across Asian/Caucasian/Black and a ratio of $0.5/0.5$ across Male/Female. For each new sampled face, we assign its UV texture by choosing that is the closest 3D face in the same ethnicity and gender from existing 482 subjects.

    \textbf{Synthetic Rendering}. We use an off-the-shelf high quality rendering engine V-ray \footnote{\url{https://vray.us/}}. With artistic assistance, we set up a shader graph to render photo-realistic facial images given a custom diffuse map and a generic specular map. We manually set up 30 different lighting conditions and further randomize head rotation $[-15^{^{\circ}}, +15^{^{\circ}}]$ in roll, yaw and pitch. The background of rendered images are randomized with a large collection of indoor and outdoor images. We opt not to render eye models and mask out the eye areas when testing by using detected local eye landmarks. Please see supplementary material for more details.

\subsection{Regressing Vertex Coordinates}
\label{sec:regressing}
    Our shape regression network consists of a feature encoder and a shape decoder. Deep facial identity feature is known for its robustness under varying conditions such as lighting, head pose and facial expression, providing a naturally ideal option for the encoded feature. Although any off-the-shelf facial recognition network would be sufficient for our task, we propose to adopt Light CNN-29V2 \cite{wu2018light} due to its good balance between network size and encoding efficiency. A pre-trained Light CNN-29V2 model is used to encode an input image into a 256-dimensional feature vector. We have used a weighted per-vertex L1 loss: weight of 5 for vertices on the facial area (within a radius of 95mm from the nose tip) and weight of 1 for other vertices.
    
    For shape decoder, we have used three fully connected (FC) layers, with the output size of 128, 200 and 8,775 respectively. The last FC layer directly predicts concatenated vertex coordinates of a generic head model consisting of 2,925 points, and it is initialized with 200 pre-computed PCA components explaining more than 99\% of the variance observed in the 10,000 augmented 3D facial shapes. When compared with unsupervised learning \cite{genova2018unsupervised}, our accessibility to a high-quality prioritized 3D face scan dataset makes it possible to achieve higher accuracy by supervision.

\subsection{Non-rigid Deformation}
\label{sec:nonrigid}
    3D vertex coordinates generated by the shape regression neural network is not directly applicable to texture projection because facial images usually contain unknown factors such as camera intrinsic, head pose and facial expression. Meanwhile, since shape regression predicts the overall facial shape, local parts such as eyes, nose and mouth are not accurately reconstructed; but they are equally important to quality perception when comparing against the original face image. We propose to utilize facial landmarks detected in a coarse-to-fine fashion and formulate non-rigid deformation as an optimization problem that jointly optimizes over camera intrinsic, camera extrinsic, facial expression and a per-vertex corrective field.

    \textbf{Problem Formulation}. To handle facial expressions, we transfer the expression blendshape model in FaceWarehouse \cite{Cao2014_FWH} to the same head topology with artist's assistance as $\{\mathbf{B}_1, \mathbf{B}_2, ..., \mathbf{B}_M\}$. In addition, we introduce a per-vertex correction field $\delta\mathbf{P}$ to cover out of space non-rigid deformation. Finally, a 3D face is reconstructed as $\mathbf{P}_F=\mathbf{P}+\sum\limits_{i=1}^{M}\boldsymbol{\beta}_{i}\mathbf{B}_{i}+\delta\mathbf{P}$. Camera extrinsic $\mathbf{T}$ transforms the face from its canonical reference coordinate system to the camera coordinate system. It has a 3-DoF vector $\mathbf{t}$ for translation and a 3-DoF quaternion representation $\mathbf{q}$ for rotation. Camera intrinsic $\mathbf{K}$ projects the 3D model to the image plane. During the optimization, we have found that using a scale factor $f_s$ to update the intrinsic matrix by $K=\bigl( \begin{smallmatrix}f_sf & 0 & c_x\\ 0 & f_sf & c_y\\0 & 0 & 1\end{smallmatrix}\bigr)$ leads to the best numerical stability. Here $[f, c_x, c_y]$ are all initialized from the size of the input image as $c_x=\frac{im_h}{2}$, $c_y=\frac{im_w}{2}$, and $f=max(im_h, im_w)$. Putting things together, we can represent the overall parameterized vector by $\mathbf{p}=[\boldsymbol{\beta}, 
    \delta\mathbf{P}, \mathbf{t}, \mathbf{q}, f_s]$.
    
    \textbf{Landmark Term.}
    We employ a global-to-local method for facial landmark localization. For global inference, we first detect the standard 68 facial landmarks, and use this initial estimation to crop local areas including eyes, nose, and mouth - \ie, a total of 4 cropped images. Then we perform fine-scale local inference on the cropped images (Please see the supplementary material for more details). The landmark localization approach produces a set of facial landmarks $\mathbf{L}$ where $\mathbf{L}_i=[L_i^x, L_i^y]$. We propose to minimize the distance between the predicted landmarks on the 3D model and the detected landmarks,
    
    \begin{equation}
        \label{eq:landmark}
        E_{l}=\frac{100}{W_{eye}}\sum\limits_{i=1}^{K}\|\mathcal{K}(\mathcal{T}(\mathcal{S}_{\mathcal{M}}(\mathbf{P}_F, \mathbf{m}_i), \mathbf{T}),\mathbf{K})-\mathbf{L}^i\|^2,
    \end{equation}
    where $\mathcal{S}_{\mathcal{M}}(\mathbf{P}, \mathbf{m}_i)$ samples a 3D vertex from $\mathbf{P}$ given a production-ready and sparse triangulation $\mathcal{M}$ on barycentric coordinates $\mathbf{m}_i$, $\mathcal{K}(\cdot, \cdot)$ and $\mathcal{T}(\cdot, \cdot)$ are perspective projection and rigid transformation operators respectively, $W_{eye}$ is the distance between two outermost eye landmarks and $\frac{100}{W_{eye}}$ is used to normalize the eye distance to 100. We pre-select $\mathbf{m}$ on $\mathcal{M}$ and follow the sliding scheme \cite{cao2014displaced} to update the barycentric coordinates of the 17 facial contour landmarks at each iteration.  
    
    \textbf{Corrective Field Regularization}.
    To enforce a smooth and small per-vertex corrective field, we combine the following two losses,
    
    \begin{equation}
        \label{eq:correctivefield}
        E_{c}=\|\mathcal{L}(\mathbf{P}_F, \mathbf{M})-\mathcal{L}(\mathbf{P}+\sum\limits_{i=1}^{M}\boldsymbol{\beta}^{t-1}_{i}\mathbf{B}_{i}, \mathbf{M})\|^2+\lambda_\delta\|\delta\mathbf{P}\|^2.
    \end{equation}
    
    The first loss is used to regularize a smooth deformation by maintaining the Laplacian operator $\mathcal{L}$ on the deformed mesh (please refer to \cite{sorkine2004laplacian} for more details). $\boldsymbol{\beta}^{t-i}$ indicates the estimated facial expression blendshape weights from the last iteration and is a fixed value. The second loss is used to enforce a small corrective field and $\lambda_\delta$ is used to balance the two terms.
    
    \textbf{Other Regularization Terms}.
    We further regularize on facial expression, focal length scale factor, and rotation component of camera extrinsic as follows,
    
    \begin{equation}
        \label{eq:regularization}
        E_{r}=\sum\limits_{i=1}^{M}\frac{\boldsymbol{\beta}^2_{i}}{\boldsymbol{\sigma}^2_{i}}+\lambda_f\log^2(f)+\lambda_q\|\mathbf{q}\|^2,
    \end{equation}
    where $\boldsymbol{\sigma}$ is the vector of eigenvalues of the facial expression covariance matrix obtained via PCA. $\lambda_f$ and $\lambda_q$ are regularization parameters.
    
    \textbf{Summary}. Our total loss function is given by
    \begin{equation}
        \label{eq:total}
        E=E_{l}+\omega_cE_{c}+\omega_rE_{r},
    \end{equation}
    where $\omega_c$ and $\omega_r$ are used to balance relative importance of the three terms. $E$ is optimized by Gauss-Newton approach over parameters $\mathbf{p}^t$ for a total of $N$ iterations. For the initial parameter vector $\mathbf{p}^0$, $\boldsymbol{\beta}^0$ and $\delta\mathbf{P}^0$ are initialized as all-$0$ vectors, $\mathbf{t}^0$ and $\mathbf{q}^0$ are estimated from the EPnP approach \cite{lepetit2009epnp}, and $f_s^0$ is initialized to be $1$.

\subsection{Texture Processing}
\label{sec:texture}

    Upon non-rigid deformation, we project selfie texture to the UV space of the generic model using the estimated camera intrinsic, head pose, facial expression and per-vertex correction. While usually only the frontal area on a selfie is visible, we recover textures on other areas, \eg, back of head and neck, by using the UV texture of one of the 482 subjects that is \emph{closest} to the query subject. We define closeness as L1 loss on the distance between LightCNN-29V2 embeddings, \ie, through face recognition. Finally given a foreground projected texture and a background default texture, we blend them using the Poisson Image Editing \cite{perez2003poisson}.

%% file: experimental_results.tex
\section{Experimental Results}
\label{sec:experiments}

\subsection{Implementation Details}

    For shape regression, we use Adam optimizer with a learning rate of 0.0001 and the momentum $\beta_1=0.5$, $\beta_2=0.999$ for 500 epochs. We train on a total of 10,000 synthetically rendered facial images with a batch size of 64. For non-rigid deformation, we use a total of $N=5$ iterations. When minimizing Eq.~\eqref{eq:total}, we use $\omega_c=25$ and $\omega_r=10$. In Eq.~\eqref{eq:correctivefield}, we set $\lambda_\delta=4$, and in Eq.~\eqref{eq:regularization} we set $\lambda_f=5$ and $\lambda_q=5$.

\subsection{Database and Evaluation Setup}
    
    \textbf{Stirling/ ESRC 3D Faces Database} The ESRC \cite{Feng2018esrc} is the latest public 3D faces database captured by a Di3D camera system. The database also provides several images captured from different viewpoints under various lighting condition. We select those subjects who have both 3D scan and a frontal neutral face for evaluation. There are total 129 subjects (62 male and 67 female) for testing. Note that in this dataset, around $95 \%$ of people are Caucasian. 
        
    \textbf{JNU-Validation Database} The JNU-Validation Database is a part of the JNU 3D face Database collected by the Jiangnan University \cite{Koppen2018GMM}. It has $161$ 2D images of $10$ Asians and their 3D face scans captured by 3dMD. Since the validation database was not used during training, we consider it as a test database for Asians. The 2D images of each subject are in range of $[3, 26]$. To minimize the impact of imbalance data, we select three frontal images of each subject for quantitative comparison.
        
    \textbf{Our Test Data} Since there is no public database available for testing, which shall cover all the gender and races, we randomly pick five subjects from the six group in Table \ref{tab:race} and form a total 30 subjects as the evaluation database. The other 482 scans are used as for data augmentation and training/validation stage for both geometry and texture. Each subject has two testing images: a selfie captured by a Samsung Galaxy S7 and an image captured from a Sony a7R DSLR camera by a photographer. 
    
    \textbf{Evaluation Setup} We compared our method with several state-of-the-art-methods including 3DMM-CNN \cite{tuan2017regressing}, Extreme 3D Face (E3D) \cite{tuan2018extreme}, PRNet \cite{feng2018prn}, RingNet \cite{sanyal2019learning}, and GanFit \cite{gecer2019ganfit}. The reconstructed model detail of each methods are shown in Table \ref{tab:space}.  Note that for our method and RingNet, both eyes, teeth and tongue and their model holders are removed before comparison. Because the evaluation metric is using the point-to-plane error, unrelated data will increase the over all error. Although removing those parts will also slightly increase the error (e.g., no data in the eyes area to compare), the introduced error is much smaller than the error of directly using the original models.
    
    \begin{table*}
        \center
        \begin{tabular}{|c|c|c|c|c|c|c|}
        \hline
               & Ours & RingNet\cite{sanyal2019learning} & GanFit \cite{gecer2019ganfit} &  PRNet\cite{feng2018prn} & E3D\cite{tuan2018extreme}  & 3DMM-CNN\cite{tuan2017regressing}  \\ \hline
        Full Head    & Yes  & Yes  & No      & No     & No      & No      \\ \hline
        Vertex  & 2.9K (2.7K) & 5.0K (3.8K) & 53.2K     & 43.7K  & $\sim$155K    &  47.0K  \\ \hline
        Face  & 5.8K (5.3K) & 10.0 K (7.4K) & 105.8K    & 86.9K  & $\sim$150K    &  93.3K  \\ \hline
        \end{tabular}
        \caption{The geometric complexity of our method and other method. Note that except E3D, the other methods used the same topology for their reconstructed model. The number inside the parentheses in both our method and RingNet are the details of head models after unrelated mesh removal.}
        \label{tab:space}
    \end{table*}

\subsection{Quantitative Comparison}
    
    \textbf{Evaluation Metric}: To align the reconstructed model with ground truth, we followed the step of \cite{tuan2017regressing, genova2018unsupervised, gecer2019ganfit} and the challenge \cite{Feng2018esrc}. Since the topology of each method is fixed, seven pre-selected vertex index is first used to roughly align the reconstructed model to the ground truth and then the model was further refined by iterative closest point (ICP) \cite{Amberg2007_nICP}. The position of vertex of the tip of the nose $v_t$ is chosen to be the center of the ground truth and reconstructed models. Given a threshold $d$ mm, we discard those vertex $v_i$, where $||v_i - v_t|| > d$. To evaluate the reconstructed model with ground truth, we used the Average Root Mean Square Error (ARMSE) \footnote{\url{https://codalab.lri.fr/competitions/572\#learn_the_details-evaluation}} as suggested by the $2{nd}$ 3DFAW Challenge \footnote{\url{https://3dfaw.github.io/}} , where it computes the closest point-to-mesh distance between the ground truth and predicted model and vice versa.
    
\begin{figure}
        \centering
        \subfloat[ESRC]{\includegraphics[width= 0.5 \linewidth]{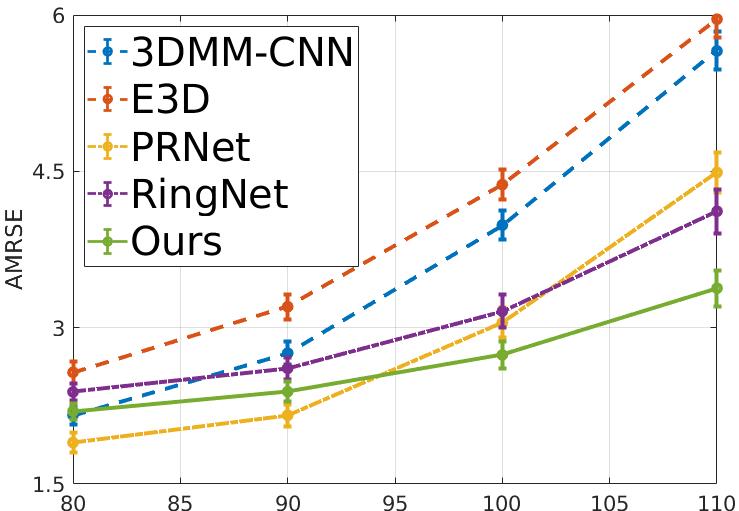}}
        \subfloat[JNU-Validation]{\includegraphics[width= 0.5 \linewidth]{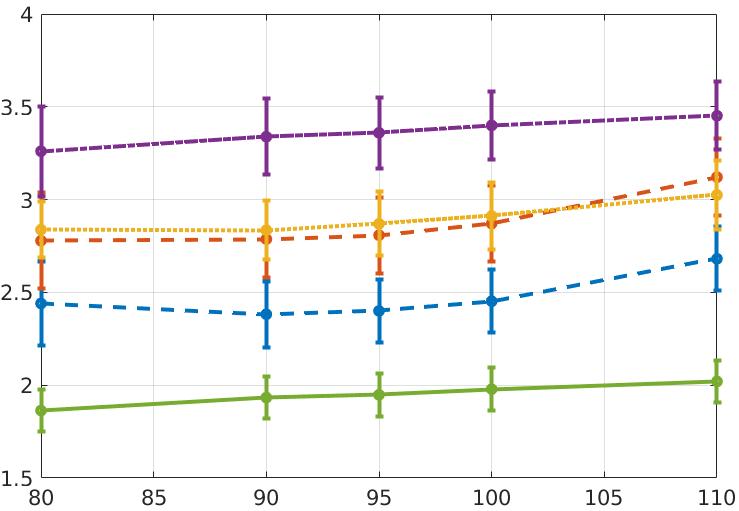}}
        \caption{The quantitative results of our method compare to 3DMM-CNN \cite{tuan2017regressing}, E3D \cite{tuan2018extreme}, PRNet \cite{feng2018prn} and RingNet \cite{sanyal2019learning} on both ESRC and JNU-Validation database.}
    \label{fig:ESRC_JNU}
    \end{figure}

    \textbf{ESRC and JNU-validation Dataset}:
    In Figure \ref{fig:ESRC_JNU}, we have chosen $d = [80, 90, 100, 110]$ and computed the ARMSE for each reconstructed model and ground truth. Note that the annotation provided by ESRC database only has the seven landmark for alignment, thus instead of using the tip of nose, we use the average of the 7 landmark as the center of face. In ESRC, our result is better than other methods when $d > 95$ and our performance is more resilient as $d$ increases. This indicates that our method can better replicate the shape of the entire head than other methods. In JNU-validation database, since other methods are trained from a Caucasian-dominated 3DMM model, while the other races are also considered during our augmented stage, we can achieve much smaller reconstructed error at every $d$ value. 
     
\begin{figure*}
        \centering
        \subfloat[]{\includegraphics[width= 0.3 \linewidth]{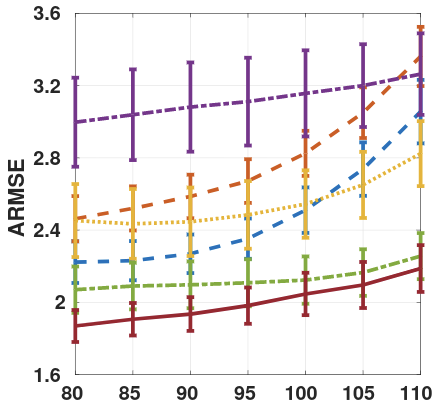}}
        \subfloat[]{\includegraphics[width= 0.5 \linewidth]{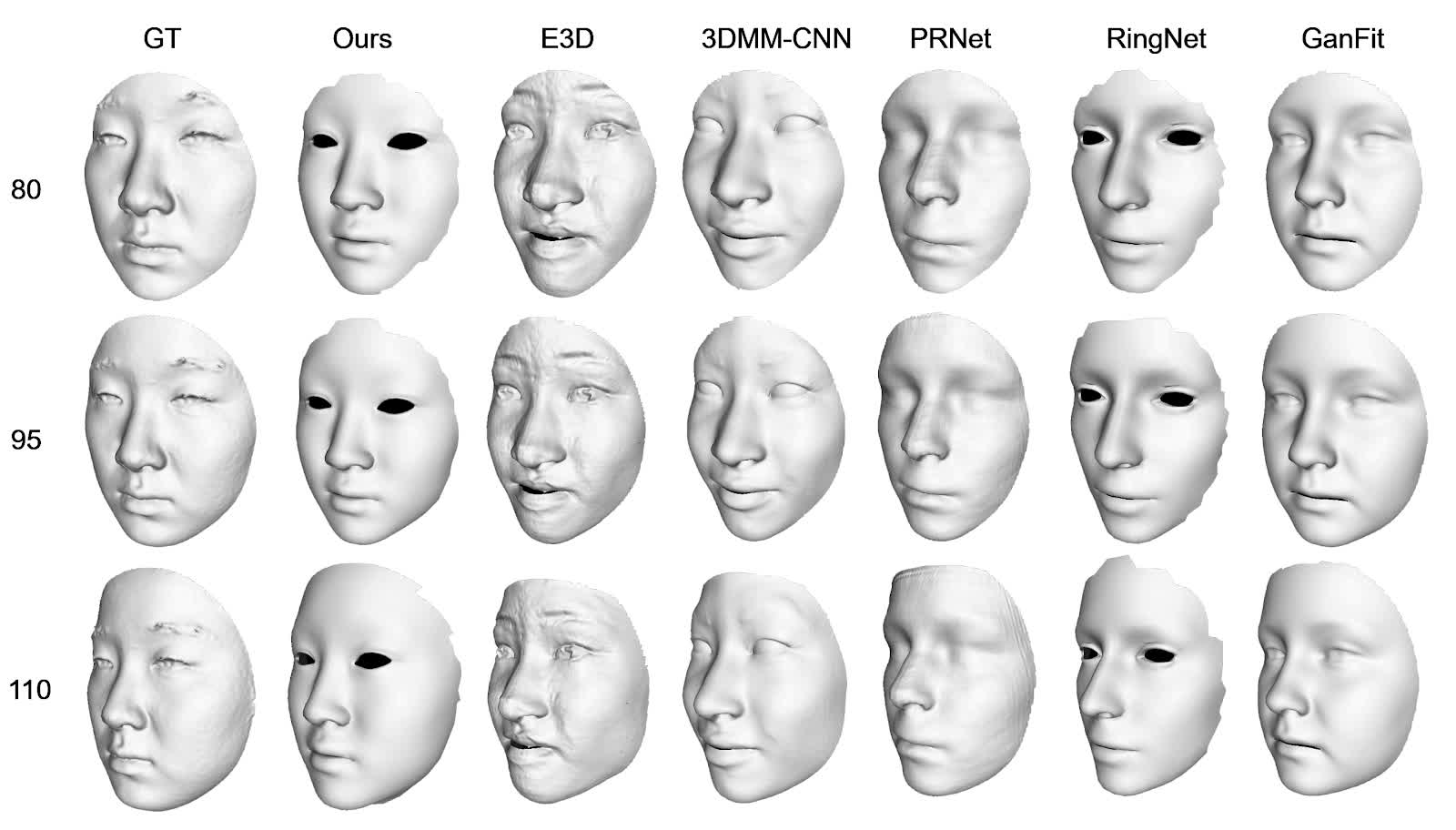}}\\
        \subfloat[ ]{\includegraphics[width= 0.8\linewidth]{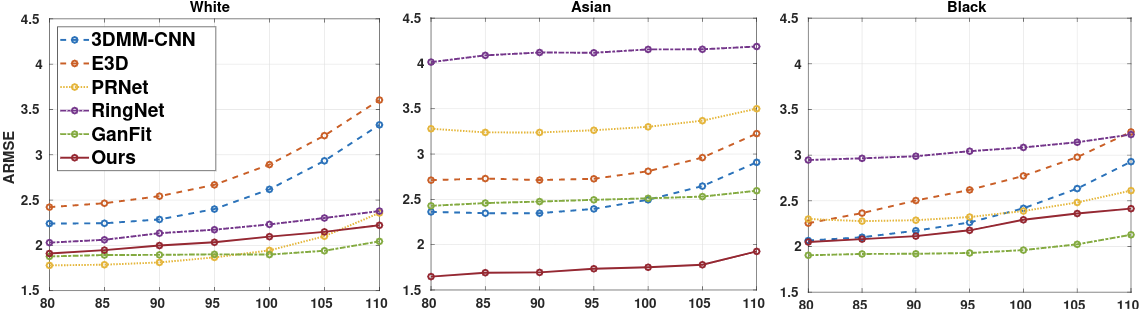}} \\
        \caption{The quantitative results of our method compare to  E3D \cite{tuan2018extreme}, 3DMM-CNN \cite{tuan2017regressing}, PRnet \cite{feng2018prn}, RingNet \cite{sanyal2019learning} and GanFit \cite{gecer2019ganfit}. (a) The overall performance of each method. (b) The qualitative comparison of cropped meshes with ground truth in $d = [80, 95, 110]$. (c) The evaluation results on different ethnicity.}
        \label{fig:quantitative_class}
    \end{figure*}

\begin{figure}
        \centering
        \includegraphics[width= 1.0 \linewidth]{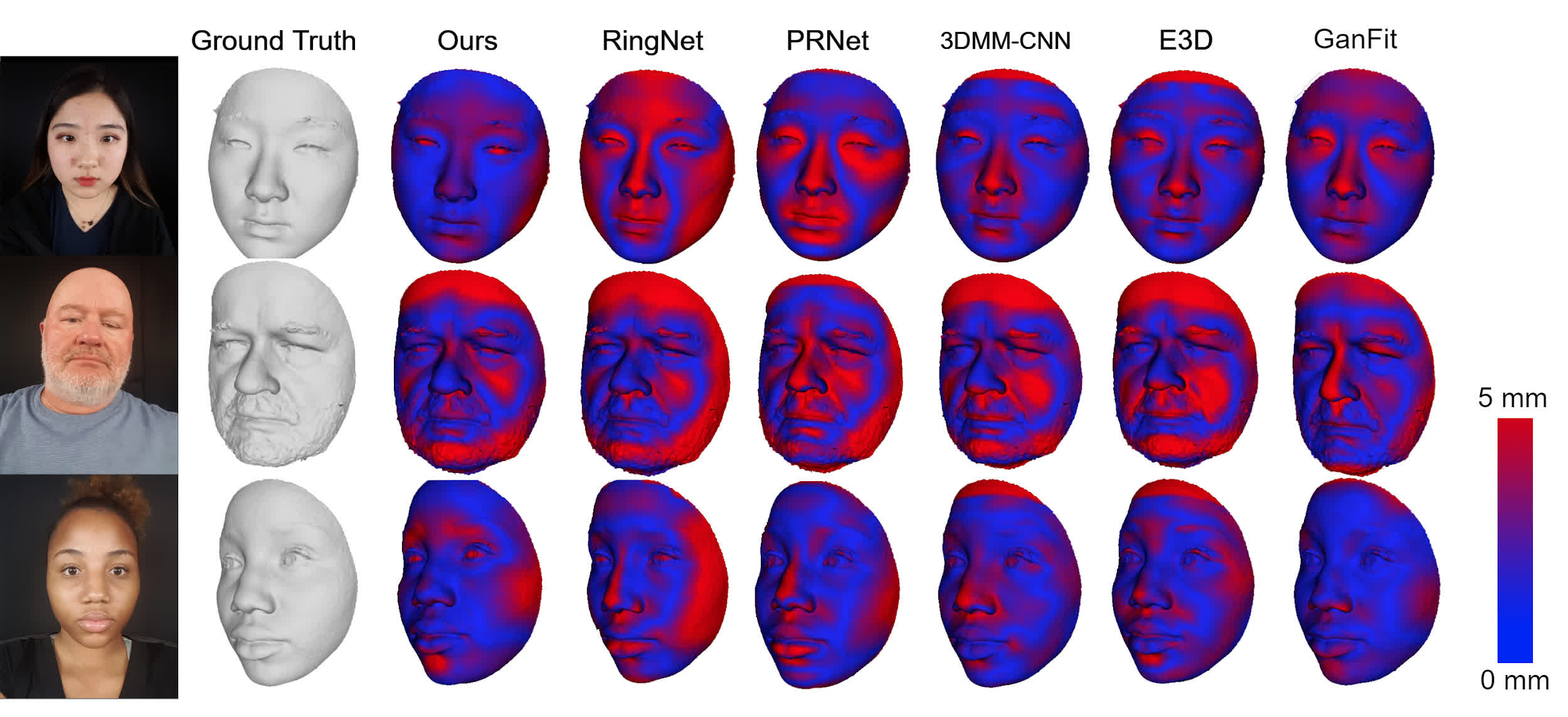}
        \caption{The heatmap visualization of reconstructed models in $d = 110$. Vertices colored red fall above the 5mm error tolerance, while blue vertices are those which lie within the tolerance. }
    \label{fig:heatmap}
    \end{figure}

    \textbf{Our Test Dataset}:
    In Figure \ref{fig:quantitative_class} (a), the centers of each error-bar are the average of the ARMSE from the 60 reconstructed meshes. The range of the errorbar is $\pm 1.96 \times SE$, where $SE$ is the standard error. It is shown that our reconstructed models is slightly better than GanFit and significantly better than other methods. It is worth mentioning that our vertex number is only $\sim 70\%$ of RingNet and less than $6\%$ of other methods. In Figure \ref{fig:quantitative_class} (b), the cropped mesh of the ground truth and each methods are shown under different threshold of $d$. To utilize the reconstructed models for real-world application, we believe that $d = 110$ is the best value because it captured the entire head instead of the frontal face. We further investigate the performance under different races and the results are shown in Figure \ref{fig:quantitative_class} (c). Our method can correctly replicate the model to under $2.5$ mm of error in all ethnicity, while other methods such as RingNet and PRNet are very sensitive to the ethnicity differences. Although GanFit performed slightly better than our method on White and Black races, the overall performance is not as good as ours because they are not able to recover the Asian geometries well. It is worth noting that we used 10000 synthetic images augmented from less than 500 scan data, which is only 5\% of the data used in GanFit. To fairly visualize the error between methods without the effect of different topology, we find the closest point-to-plane distance from ground truth to reconstructed model and generate the heat-map for each method in Figure \ref{fig:heatmap}.

\subsection{Ablation Study}
    To demonstrate the effectiveness of the individual modules in the proposed approach, we modify one variable at a time and compare with the following alternatives:
        
    $\bullet$ \textbf{No Augmentation (No-Aug)}: Without any augmentation, we simply repetitively sample 10,000 faces from 482 subjects.
    
    $\bullet$ \textbf{Categorized-PCA Sampling (C-PCA)}: Instead of DR Feature based sampling, we propose a PCA based sampling method. We train a shape PCA model from 482 subjects, and for each group in Table \ref{tab:race}, a Gaussian random vector $\mathbf{x} \sim \mathcal{N}(\mu_i,\,\Sigma_i^{2})$ is used to create weights of the principal shape components, where $\mu_i$ and $\Sigma_i^{2}$ are the mean vector and co-variance matrix of those coefficients in the group. We sample 10,000 faces with this augmentation approach.
    
    $\bullet$ \textbf{Game engine Rendering-Unity}: Instead of using high-quality photo-realistic renderer, we use Unity, a standard game rendering engine, to synthesize facial images. The quality of rendered images are comparatively lower than V-ray. We keep the DR feature based augmentation approach and rendered exactly the same 10000 synthetic faces mentioned in Section \ref{sec:photorealistic}.
        
\begin{figure}
        \centering
        \includegraphics[width= 0.8 \linewidth]{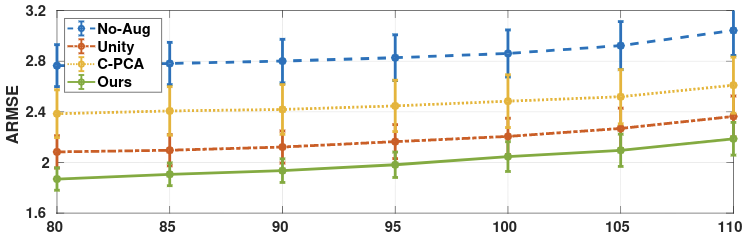}
        \caption{The quantitative results of No-Aug, C-PCA, Unity and our method. The proposed method achieve the best performance at all time.}
        \label{fig:sampling}
    \end{figure}

    In Fig. \ref{fig:sampling}, our proposed approach outperforms all other alternatives. It is expected that without data augmentation (i.e., No-Aug), the reconstructed error is the worst among all methods. The difference between C-PCA and our method proves that DR sampling augmentation creates more natural synthetic faces for training. The results between Unity and our method shows that the quality of rendered images plays an important role in bridging the gap between real and synthetic images. 

\subsubsection{Qualitative Comparison}
\begin{figure}
        \centering
        \includegraphics[width= 1.0 \linewidth]{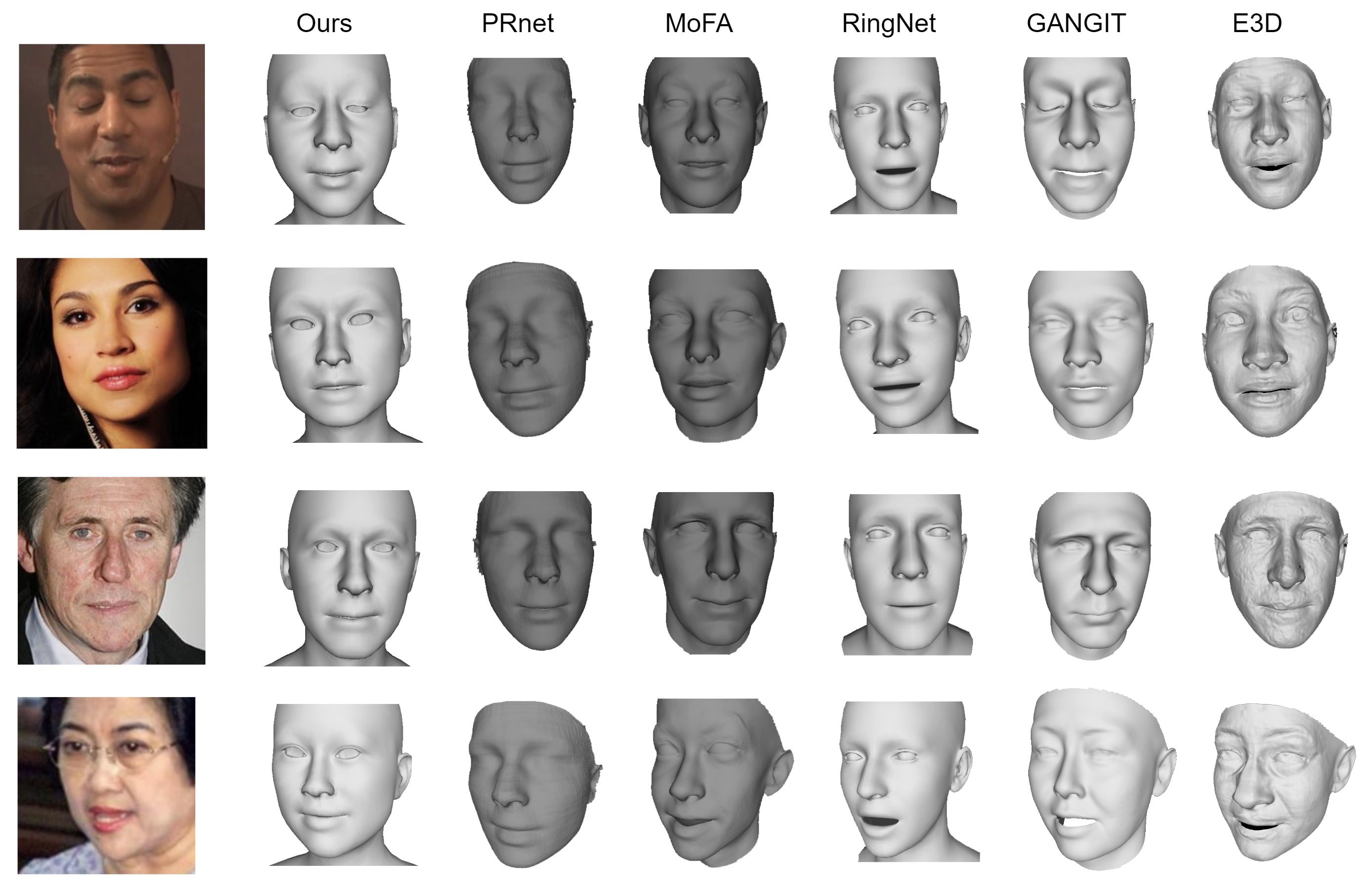}
        \caption{The qualitative comparison of our method with PRNet \cite{feng2018prn}, MoFA \cite{Tewari2017MoFAMD}, RingNet \cite{sanyal2019learning}, GanFit \cite{gecer2019ganfit} and E3D \cite{tuan2018extreme}. Our method accurately reconstructs the geometry, while maintaining a much lower vertices count, which is more suitable for production.}
    \label{fig:qualitavive_01}
    \end{figure}

\begin{figure}
        \centering
        \includegraphics[width= 0.9 \linewidth]{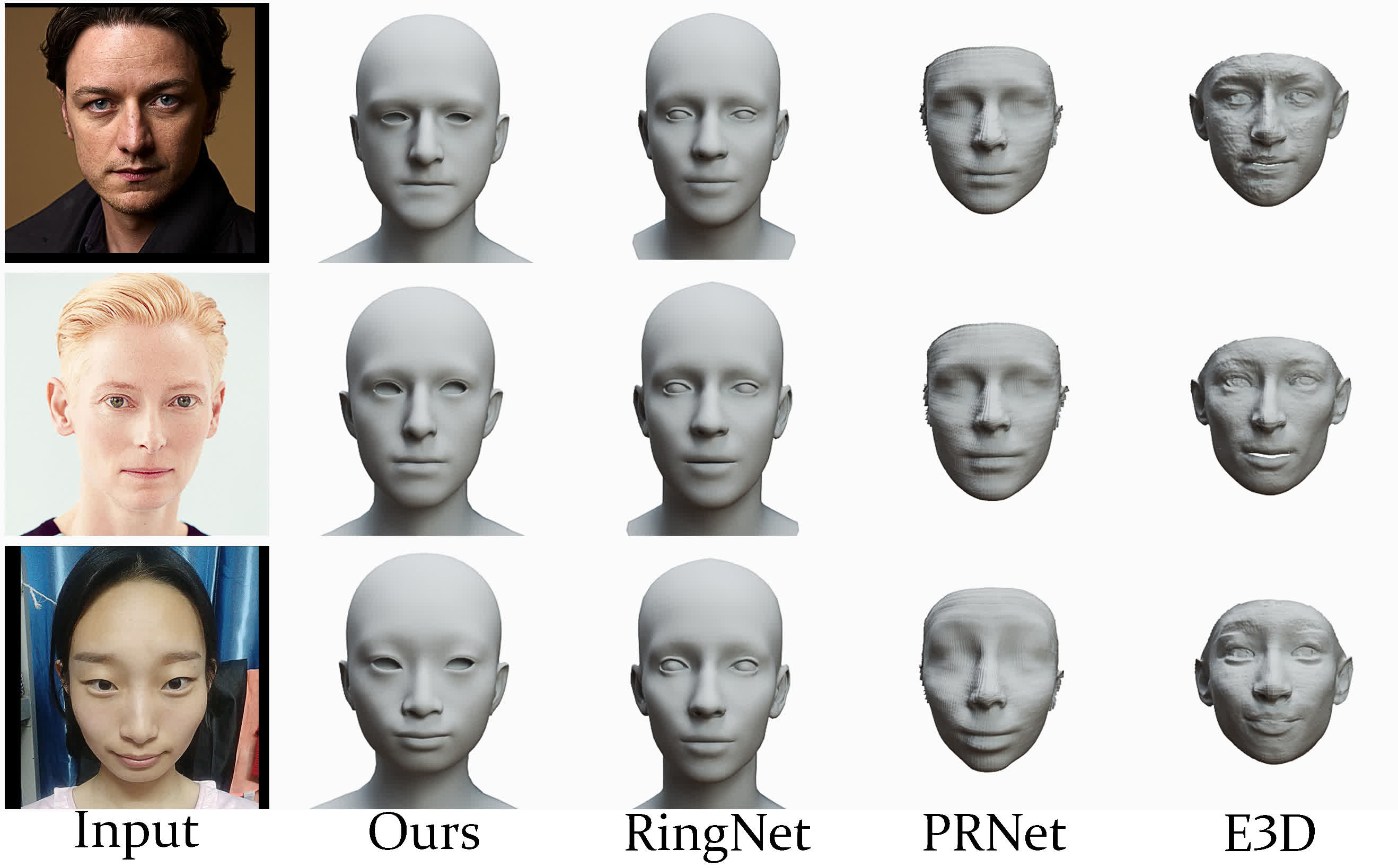}
        \caption{The showcase of our reconstruction results of several celebrities comparing to RingNet \cite{sanyal2019learning}, PRNet \cite{feng2018prn} and E3D \cite{tuan2018extreme}.}
    \label{fig:qualitavive_02}
    \end{figure}

\begin{figure}
        \centering
        \includegraphics[width= 0.75 \linewidth]{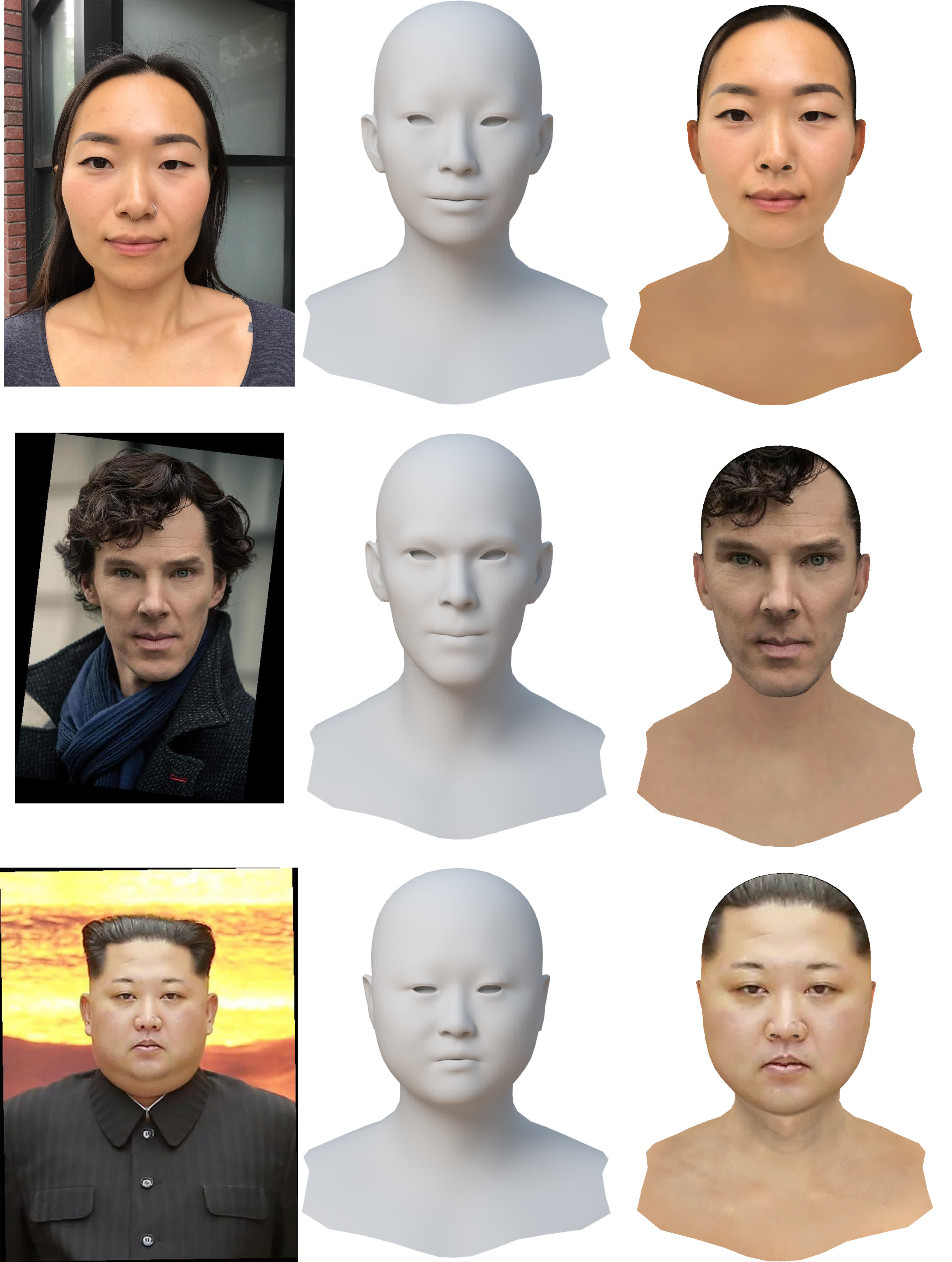}
        \caption{Our final results with blended diffuse maps. }
    \label{fig:texture}
    \end{figure}

     Figure \ref{fig:qualitavive_01} shows our shape estimation method on frontal face images side-by-side with the state-of-the-arts in MoFA test database. We picked the same images shown in GanFit \cite{gecer2019ganfit}. Our method creates accurate face geometry, while also capturing discriminate features which allow the identity of each face to be easily distinguishable from the others. Meanwhile, as shown in Table \ref{tab:space}, our result maintains a low geometric complexity. This allows our avatars to be production ready even in demanding cases such as on mobile platforms. In Figure \ref{fig:qualitavive_02}, we choose a few celebrity to verify the geometry accuracy of our method comparing to others. In Figure \ref{fig:texture}, we demonstrate our final results with blended diffuse maps in Section \ref{sec:texture}.   
    
%% file: conclusion.tex
\section{Conclusions and Future Works}
\label{sec:conclusion}

    In this paper, we demonstrated a supervised learning approach for estimating high quality 3D face shape with photo-realistic high-resolution diffuse map. To facilitate facial image synthesis, we have collected and processed a prioritized 3D face database, from which we can sample augmented 3D face shape with UV-texture to render a large collection of photo-realistic facial images. Unlike  previous approaches, our method leverages the discriminative power of an off-the-shelf face recognition neural network trained on millions of synthesized photo-realistic facial images.
    
    We have demonstrated the transferable proficiency of the proposed method from the objective of accurate face recognition to fully reconstruct the facial geometry based on a single selfie. While training on synthetically generated facial imagery, we have observed strong generalization power when tested on real-world images. This opens up opportunities in many interesting applications including VR/AR, teleconferencing, virtual try-on, computer games, special effect, and so on.

%% file: supplemental_material.tex
\section*{Section 3.2. Scan Pre-processing}
\begin{figure}[b]
        \centering
        \includegraphics[width= 0.8 \linewidth]{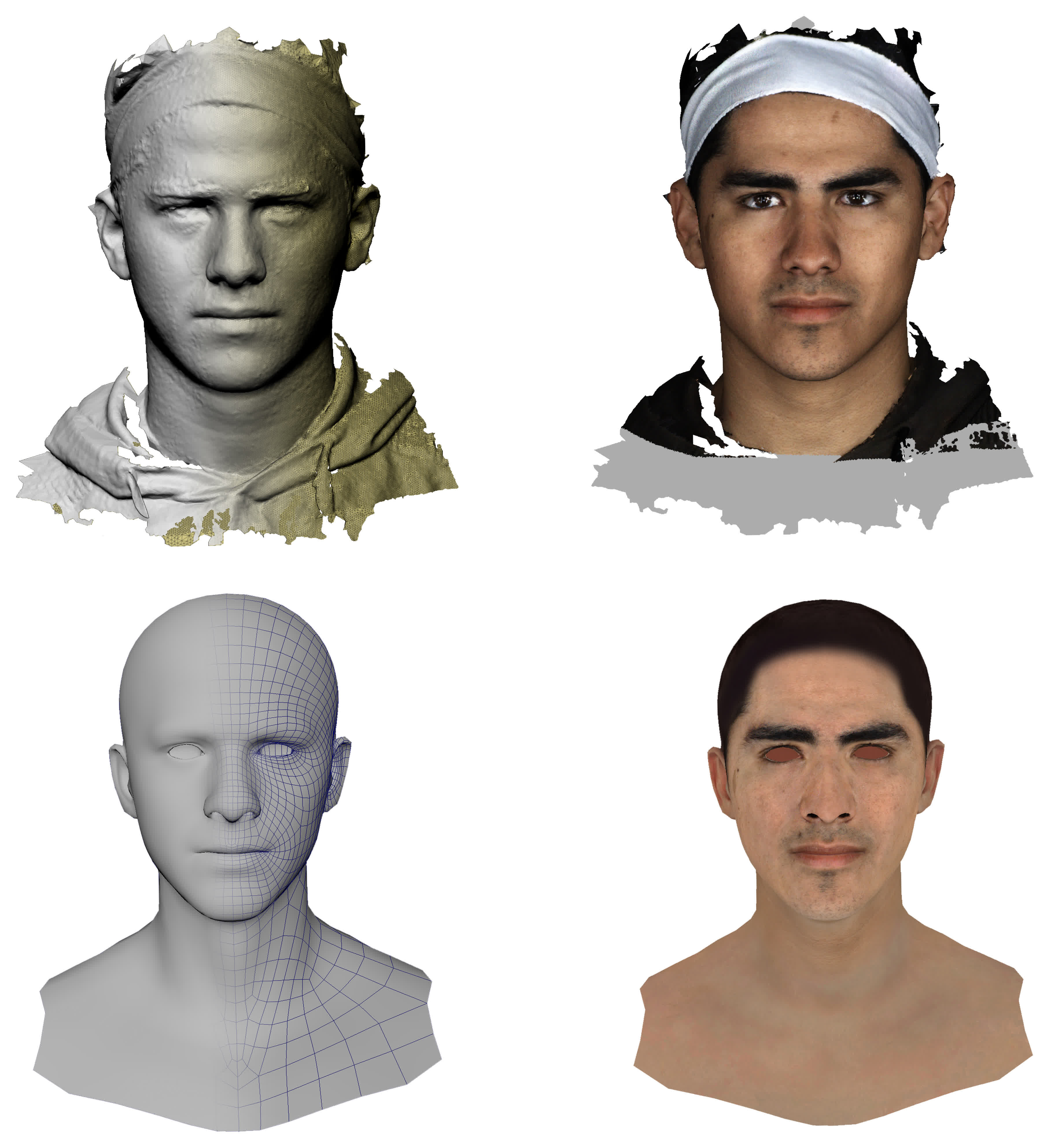}
        \caption{Top row: left is the raw facial scan with dense topology, and right is the model with UV texture; Bottom row: left is the processed face model with sparse topology, and right is the model with UV texture.}
        \label{fig:scan_model}
    \end{figure}

As shown in Fig.~\ref{fig:scan_model}, we process a raw textured 3D facial scan data to generate our 3D face representation that consists of a shape model with low polygon count and a high-resolution diffuse map for preserving details.

\section*{Section 3.2. Deformation Representation}
Here we give a detailed formulation of the Deformation Representation (DR) feature. DR feature $\mathbf{D}$ encodes local deformation around each vertex of $\mathbf{P}$ with respect to a reference mesh $\mathbf{P}^R$ into a $\mathbb{R}^9$ vector. We use the mean face of all 482 processed facial models as the reference mesh.

\textbf{Encode $\mathbf{D}$ from $\mathbf{P}$.} We denote the $i$-${th}$ vertex as $\mathbf{p}_i$ and $\mathbf{p}^R_i$ respectively. The deformation gradient in the closest neighborhood $\mathcal{N}_i$ of the $i$-${th}$ vertex from the reference model to the deformed model is defined by the affine transformation matrix $\mathbf{T}_i$ that minimizes the following energy
\begin{equation}
\label{eq:encode}
E(\mathbf{T}_i)=\sum\limits_{i\in\mathcal{N}_i}c_{ij}\|(\mathbf{p}_i-\mathbf{p}_j)-\mathbf{T}_i(\mathbf{p}^R_i-\mathbf{p}^R_j)\|^2,
\end{equation}
where $c_{ij}$ is the cotangent weight depending on the reference model to handle irregular tessellation. With polar decomposition, $\mathbf{T}_i$ is decomposed into a rotation component $\mathbf{R}_i$ and a scaling/shear component $\mathbf{S}_i$ such that $\mathbf{T}_i=\mathbf{R}_i\mathbf{S}_i$. The rotation matrix can be represented with a rotation axis $\mathbf{\omega}_i$ and rotation angle $\theta_i$ pair, and we further convert them to the matrix logarithm representation:
\begin{equation}
\label{eq:log}
log\mathbf{R}_i=\theta_i\begin{pmatrix} 
0 & -\omega_{i,z} & \omega_{i,y} \\
\omega_{i,z} & 0 & -\omega_{i,x} \\
-\omega_{i,y} & \omega_{i,x} & 0
\end{pmatrix}.
\end{equation}
Finally the DR feature for $\mathbf{p}_i$ is represented by $\mathbf{d}_i=\{log\mathbf{R}_i; \mathbf{S}_i-\mathbf{I}\}$ where $\mathbf{I}$ is the identity matrix. Since $\|\omega_i\|=1$ and $\mathbf{S}_i$ is symmetric, $\mathbf{d}_i$ has 9 DoF. 

\textbf{Recover $\mathbf{P}$ from $\mathbf{D}$.} Given the DR feature $\mathbf{D}$ and the reference mesh $\mathbf{P}^R$, we first recover the affine transformation $\mathbf{T}_i$ for each vertex. Then we try to recover the optimal $\mathbf{P}$ that minimizes:
\begin{equation}
\label{eq:decode}
E(\mathbf{P})=\sum\limits_{\mathbf{p}_i\in\mathbf{P}}\sum\limits_{j\in\mathcal{N}_i}c_{ij}\|(\mathbf{p}_i-\mathbf{p}_j)-\mathbf{T}_i(\mathbf{p}^R_i-\mathbf{p}^R_j)\|^2.
\end{equation}
For each $\mathbf{p}_i$, we obtain it by solving $\frac{\partial E(\mathbf{P})}{\partial\mathbf{p}_i}=0$ which gives
\begin{equation}
\label{eq:solve}
2\sum\limits_{j\in\mathcal{N}_i}c_{ij}(\mathbf{p}_i-\mathbf{p}_j)=\sum\limits_{j\in\mathcal{N}_i}\mathbf{T}_i(\mathbf{p}^R_i-\mathbf{p}^R_j).
\end{equation}
The resulting equations for all $\mathbf{p}_i\in\mathbf{P}$ lead to a linear system which can be written as $\mathbf{A}\mathbf{P}=\mathbf{b}$. By specifying the position of one vertex, we can get the single solution to the equation to fully recover $\mathbf{P}.$

\section*{Section 3.4. Landmark Localization}
\begin{figure}
        \centering
        \subfloat[Input Image]{\includegraphics[width= 0.475 \linewidth]{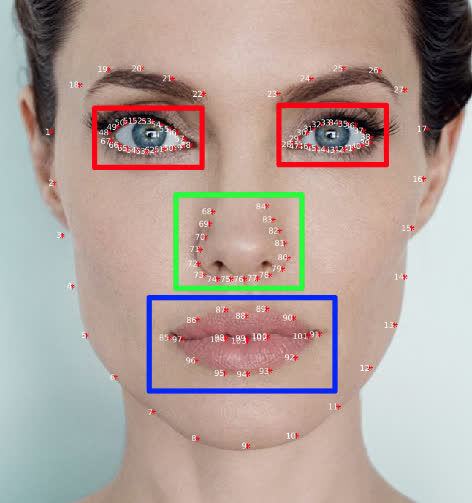}}
        \subfloat[Landmark Detection of each parts]{\includegraphics[width= 0.5 \linewidth]{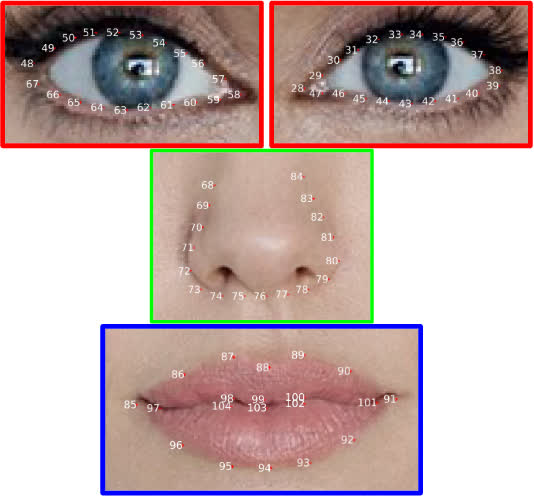}}\\
        \caption{Our landmark mark-up consists of 104 points, \ie, face contour (1-17), eye brows (18-27), left eye (28-47), right eye (48-67), nose (68-84) and mouth (85-104). (a) Coarse detection of all landmarks and corresponding bounding boxes for fine scale detection. (b) Separate fine-scale detection result of local areas.}
        \label{fig:landmark}
    \end{figure}

To achieve higher landmark localization accuracy, we have developed a coarse-to-fine approach. First, we predict all facial landmarks from the detected facial bounding box. Then, given the initial landmarks, we crop the eye, nose, and mouth areas for the second stage fine-scale landmark localization. Fig.~\ref{fig:landmark} shows our landmark mark-up as well as the bounding boxes used for the fine scale landmark localization stage. We have used a regression forest based approach \cite{kazemi2014one} as the base landmark predictor and we train 4 landmark predictors in total, \ie, for overall face, eye, nose and mouth.

\section*{Section 4.4. Different Rendering Quality}
\begin{figure*}
        \centering
        \includegraphics[width= 0.65 \linewidth]{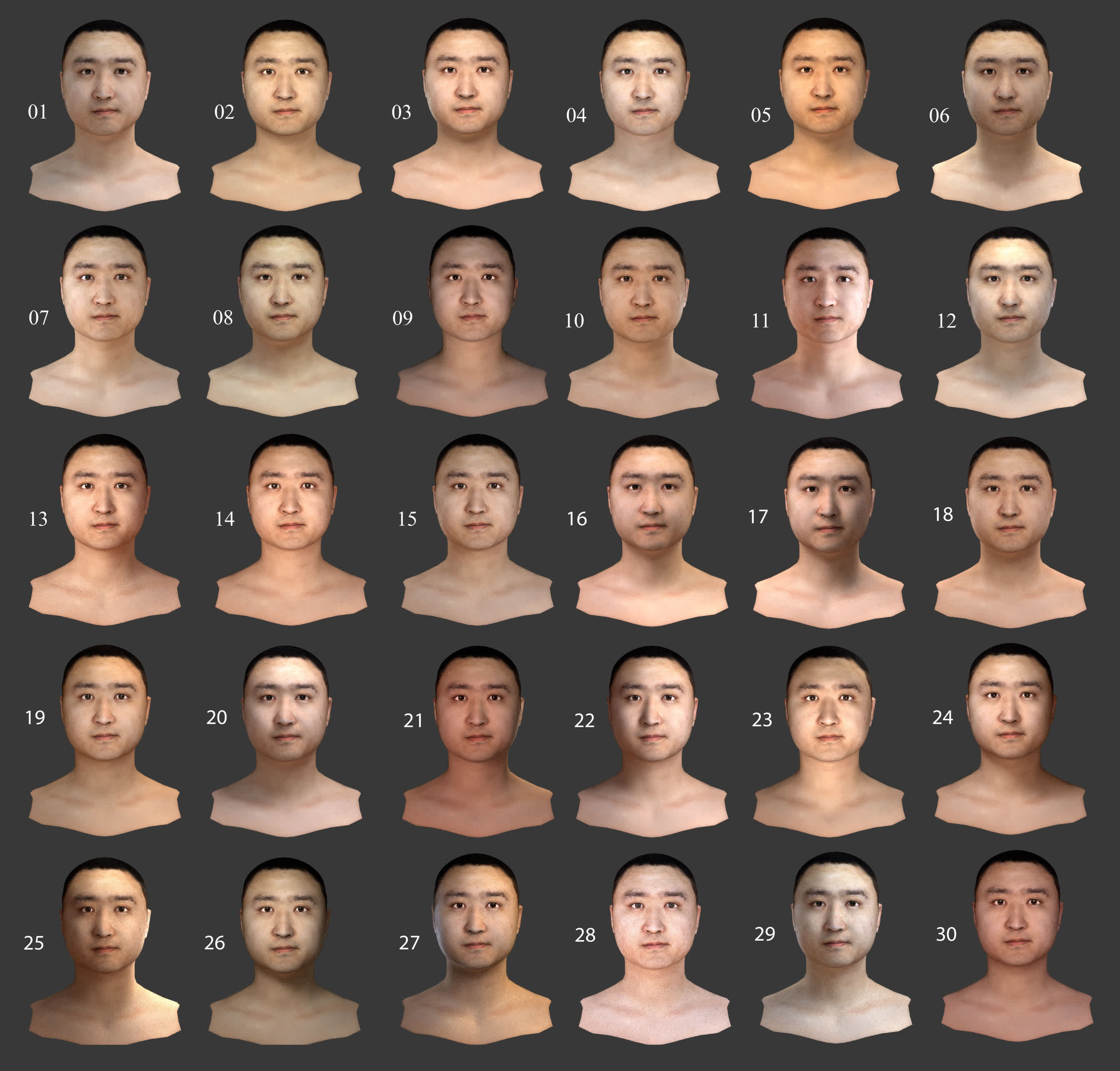}
        \caption{Different lighting conditions for photo-realistic rendering augmentation}
        \label{fig:lighting}
    \end{figure*}

\begin{figure*}
        \centering
        \subfloat[V-ray rendering samples]{\includegraphics[width= 0.65 \linewidth]{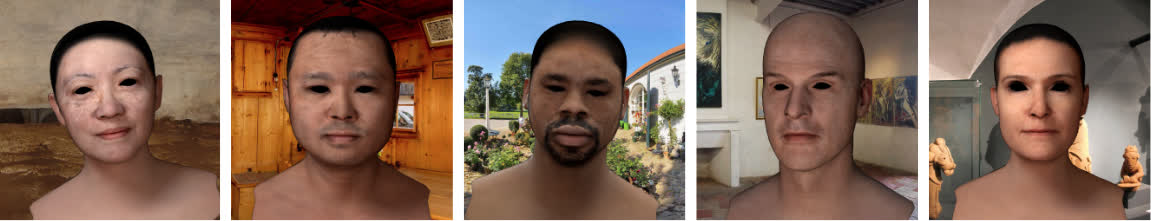}}\\
        \subfloat[Unity rendering samples]{\includegraphics[width= 0.65 \linewidth]{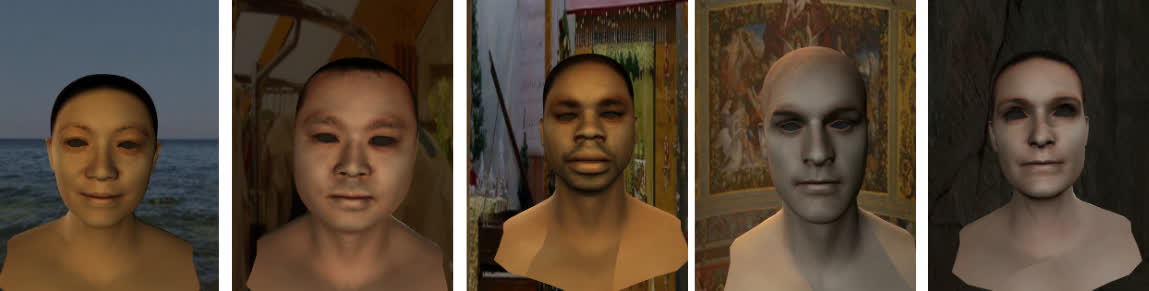}}\\
        \caption{The synthetic facial images from (a) Maya V-ray and (b) Unity}
        \label{fig:renderer}
    \end{figure*}

    In this section, first we illustrate the 30 different manually created lighting conditions used for high-quality Vray rendering as shown in Fig.~\ref{fig:lighting}. Then we provide several synthetic face images rendered from Vray and Unity as shown in Fig.~\ref{fig:renderer}. Note that, for both rendering method, we randomized the head pose, environment map, lighting condition, and the field of view (FOV) to mimic the selfie in the real world. We don't render eye models, and as a result, we mask out the eye area with detected facial landmarks during test time as mentioned in Section 3.2.

\section*{Section 4.5. More Qualitative Results}

    In this section, we provide more comparison results that cannot be included in the paper due to page limits. For GanFit \cite{gecer2019ganfit}, we have requested them to run the reconstructed results of our test data. Thus, we are only able to show the qualitative comparison with GanFit in our test database. For those images/selfies in the other database, we have compared our results with those papers whose codes are available online including RingNet \cite{sanyal2019learning}, PRNet \cite{feng2018prn}, Extreme3D \cite{tuan2018extreme} and 3DMM-CNN \cite{tuan2017regressing}.

\textbf{More Qualitative Results of Our Data}

\begin{figure*}
        \centering
        \includegraphics[width= 0.8 \linewidth]{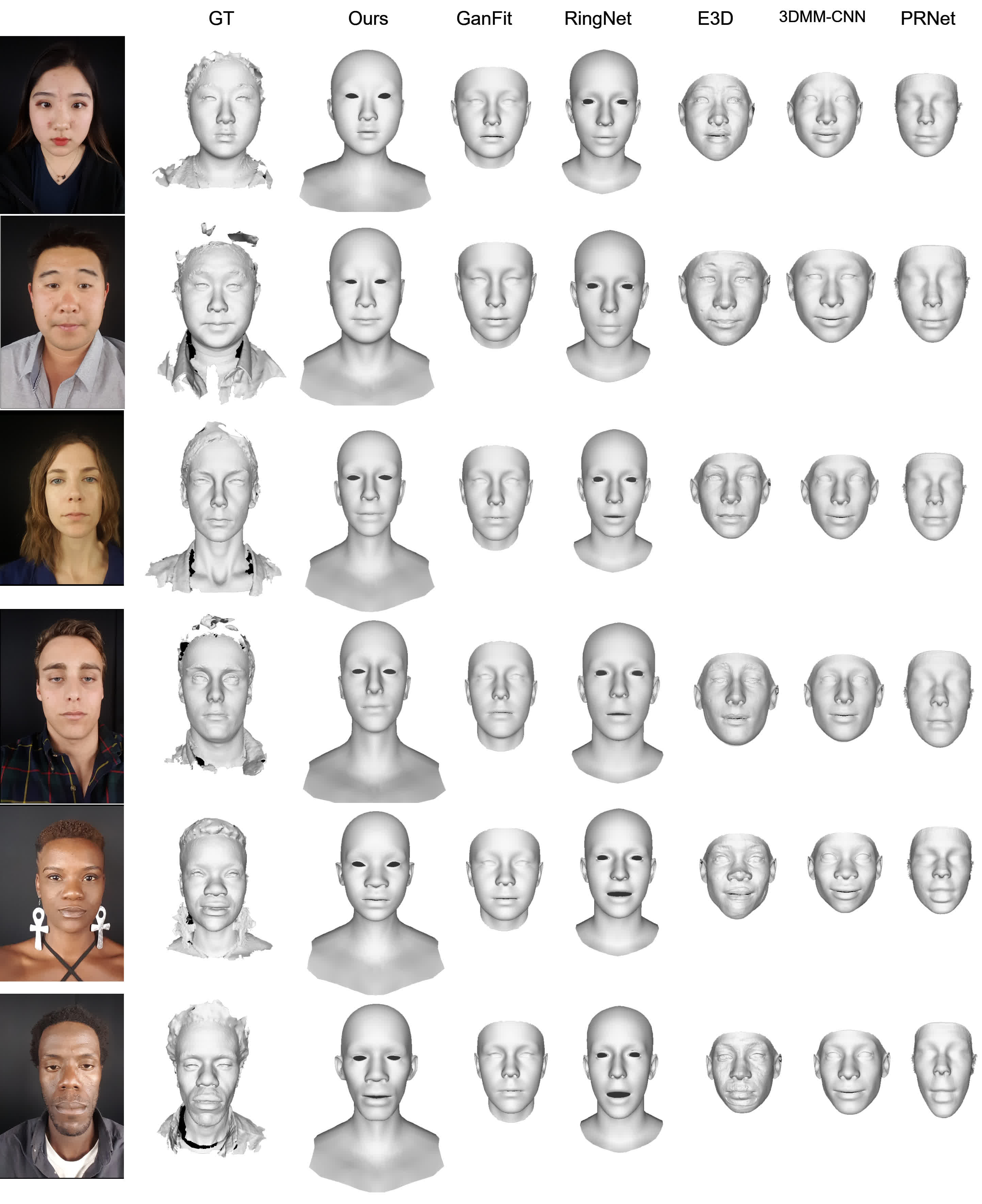}
        \caption{Qualitative results on our test dataset. From left to right, input image, ground truth, our method, GanFit \cite{gecer2019ganfit}, RingNet \cite{sanyal2019learning}, E3D \cite{tuan2018extreme}, 3DMM-CNN \cite{tuan2017regressing}, and PRnet \cite{feng2018prn}.}
        \label{fig:oben_class}
    \end{figure*}

    In Fig. \ref{fig:oben_class}, we provided the qualitative results of each categories. The first and second columns are the input image and the ground truth. Instead of showing the cropped mesh, we decided to show the whole models for each method in Fig. \ref{fig:oben_class}. It is worth noting that our reconstructed full head model is ready to be deployed for different applications.   
    
\textbf{Qualitative ESRC and JUN-Validate}

\begin{figure*}
        \centering
        \subfloat[ESRC Male]{\includegraphics[width= 0.75 \linewidth]{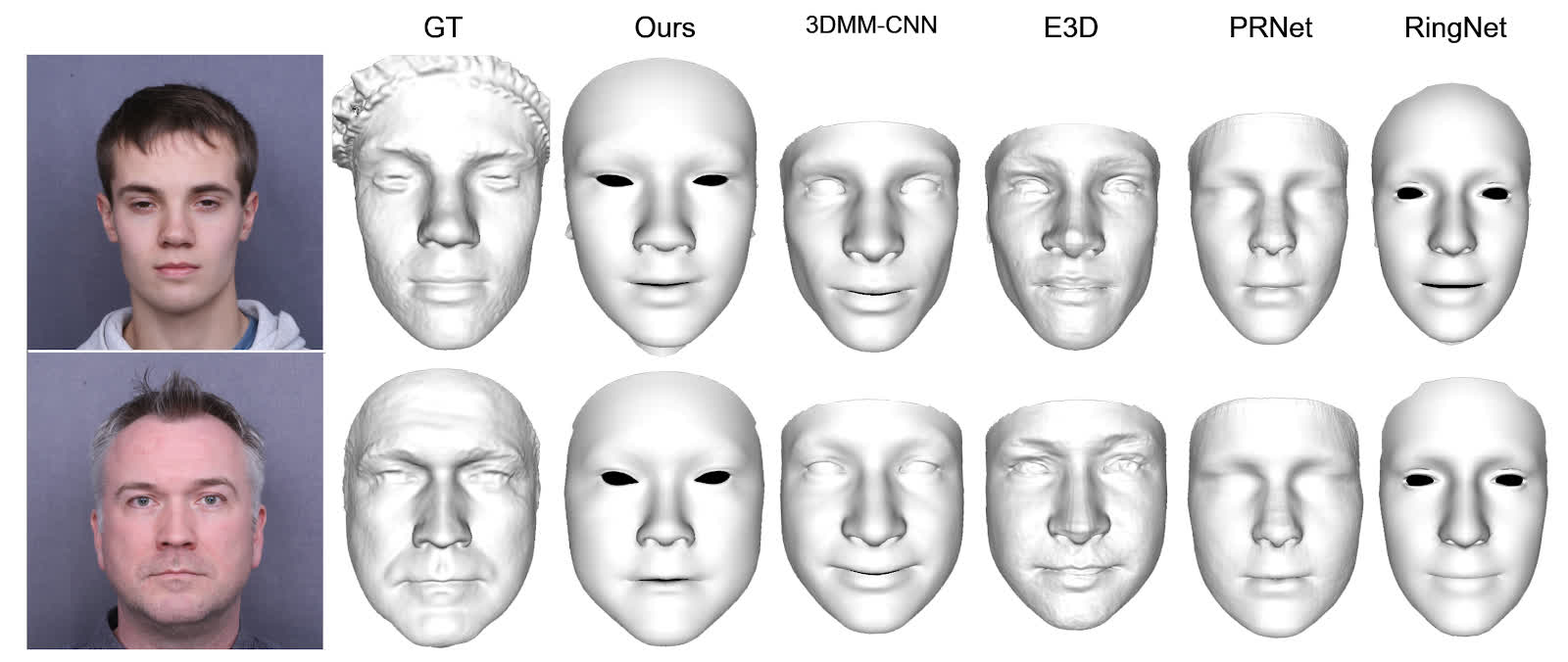}}\\
        \subfloat[ESRC Female]{\includegraphics[width= 0.75 \linewidth]{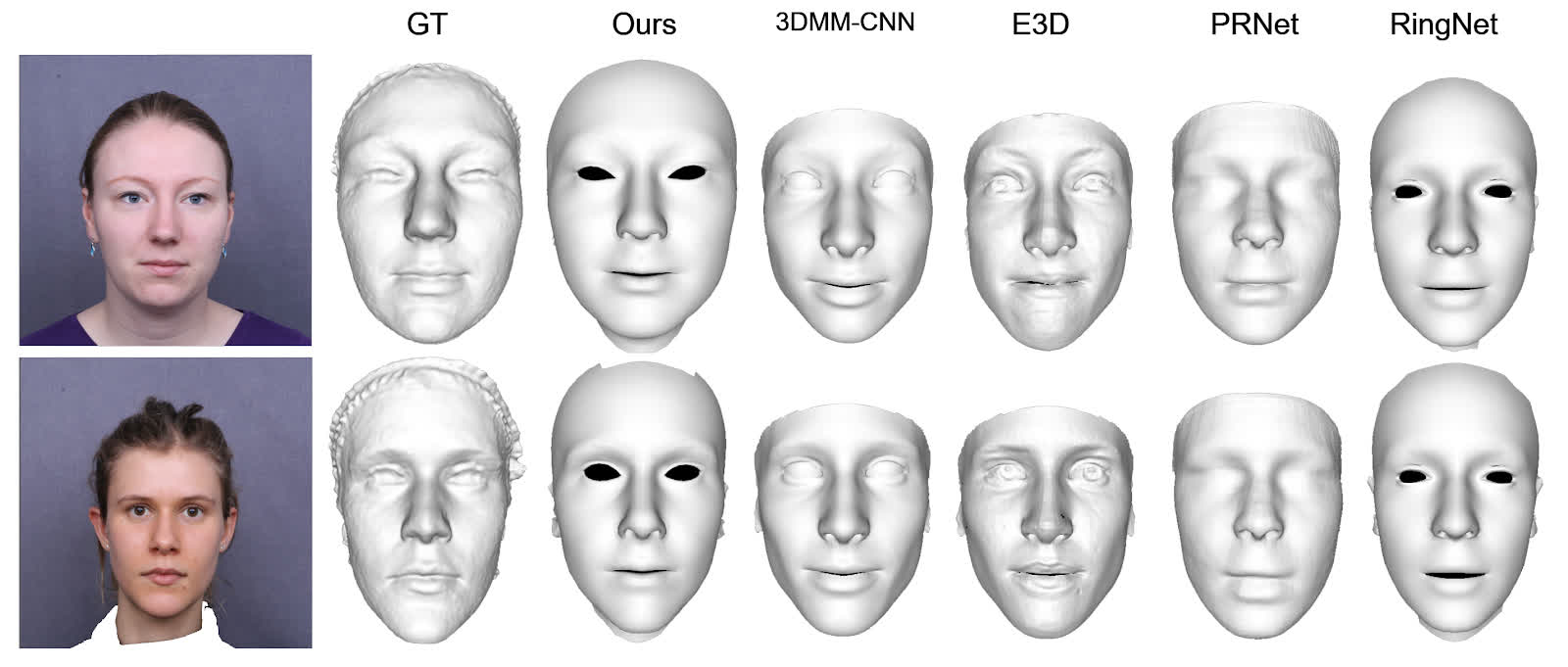}}\\
        \subfloat[JUN-Validation Database]{\includegraphics[width= 0.75 \linewidth]{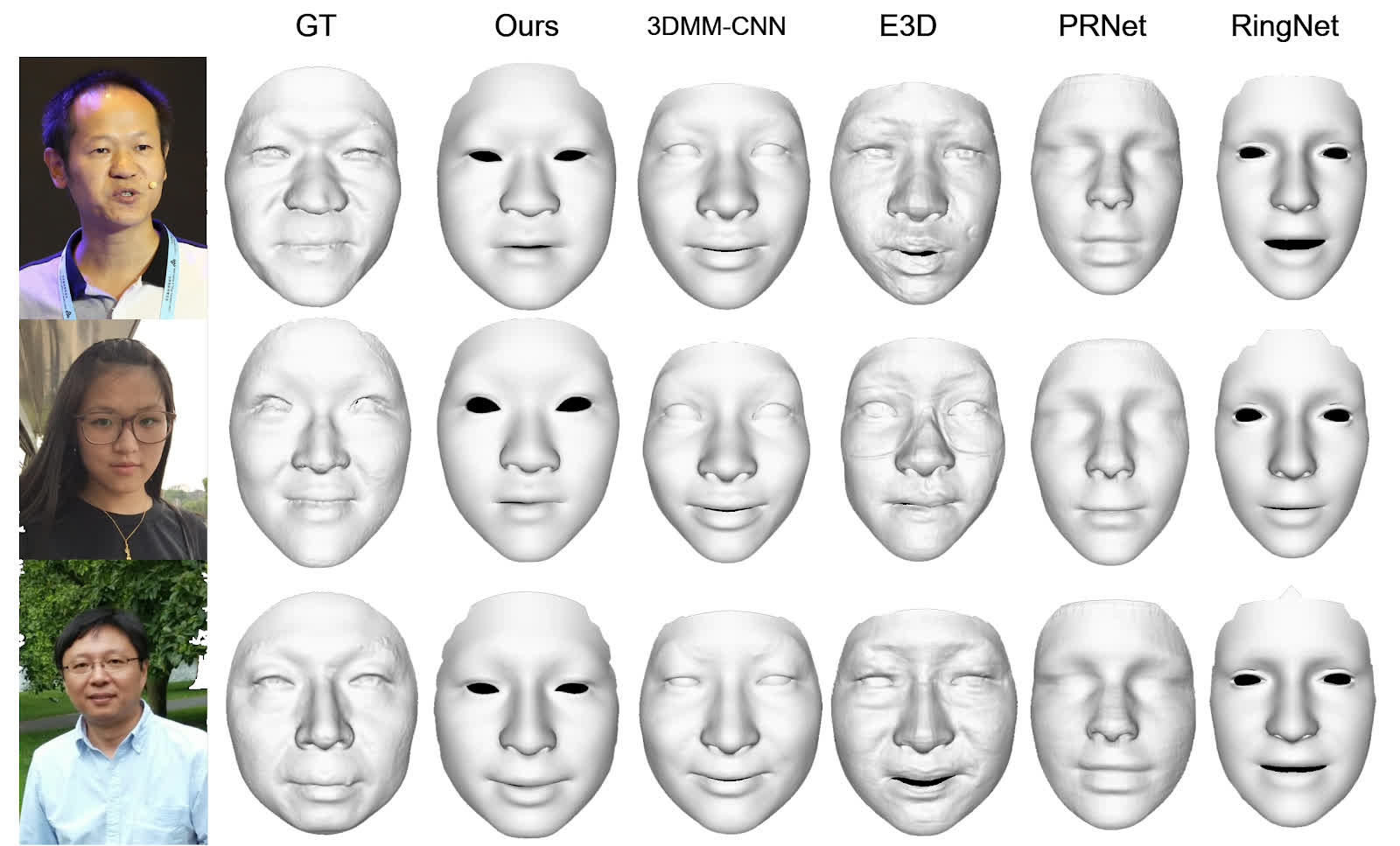}}\\
        \caption{Qualitative results on the ESRC and JNU-Validation datasets. From left to right, input image, ground truth, our method, 3DMM-CNN\cite{tuan2017regressing}, E3D \cite{tuan2018extreme}, PRnet \cite{feng2018prn}, RingNet \cite{sanyal2019learning}.}
        \label{fig:esrc_jun_qual}
    \end{figure*}

    Due to the paper limitation, we are not able to show the qualitative result of ESRC and JUN-validate Dataset. As shown in Figs. \ref{fig:esrc_jun_qual}, we can still see the similar results we claimed in the paper that the proposed method can correctly replicate the 3D models from single selfies with much lower polygon.

\begin{figure*}
        \centering
        \subfloat[MoFA Male]{\includegraphics[width= 0.8 \linewidth]{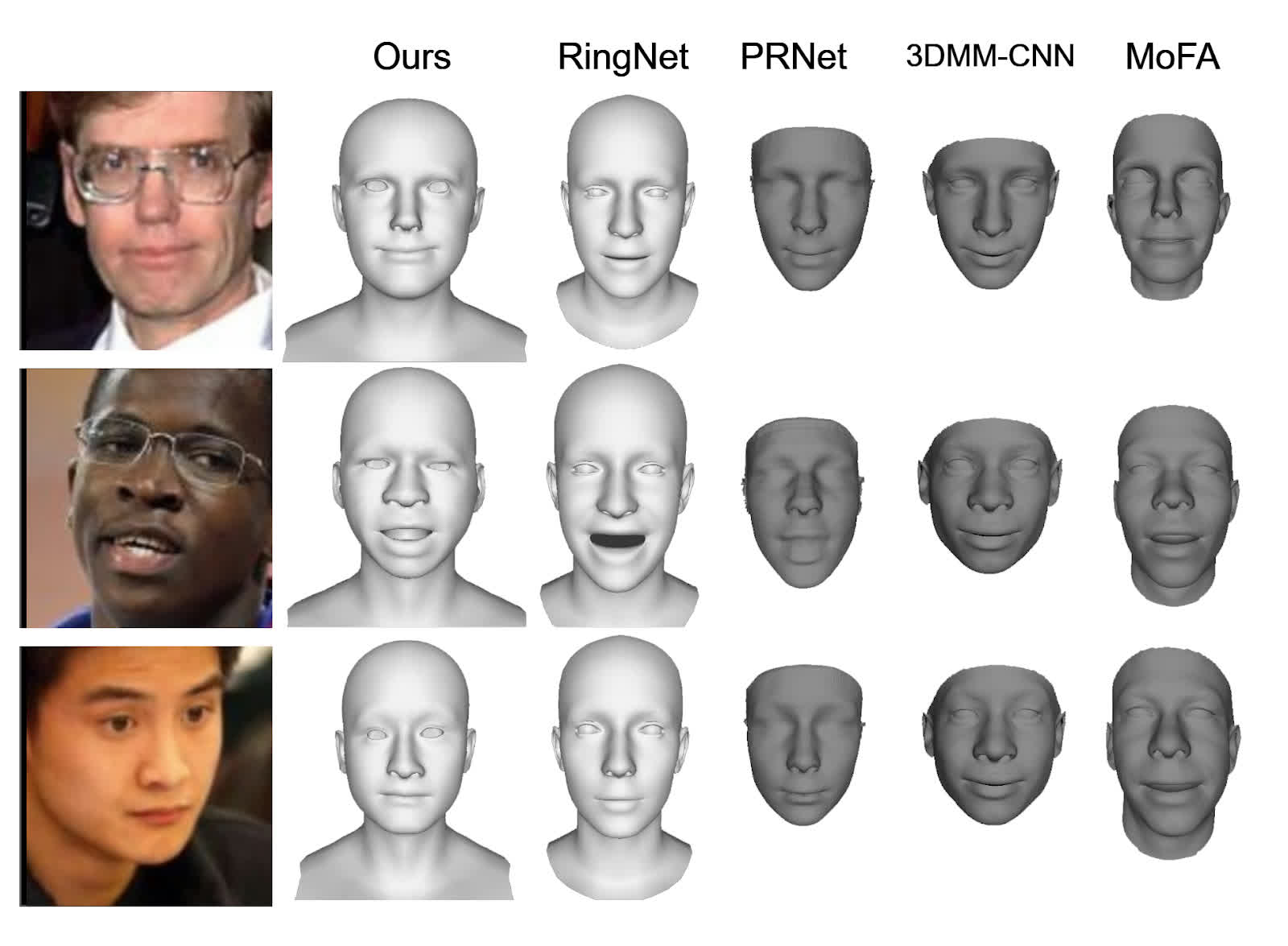}}\\
        \subfloat[MoFa Female]{\includegraphics[width= 0.8 \linewidth]{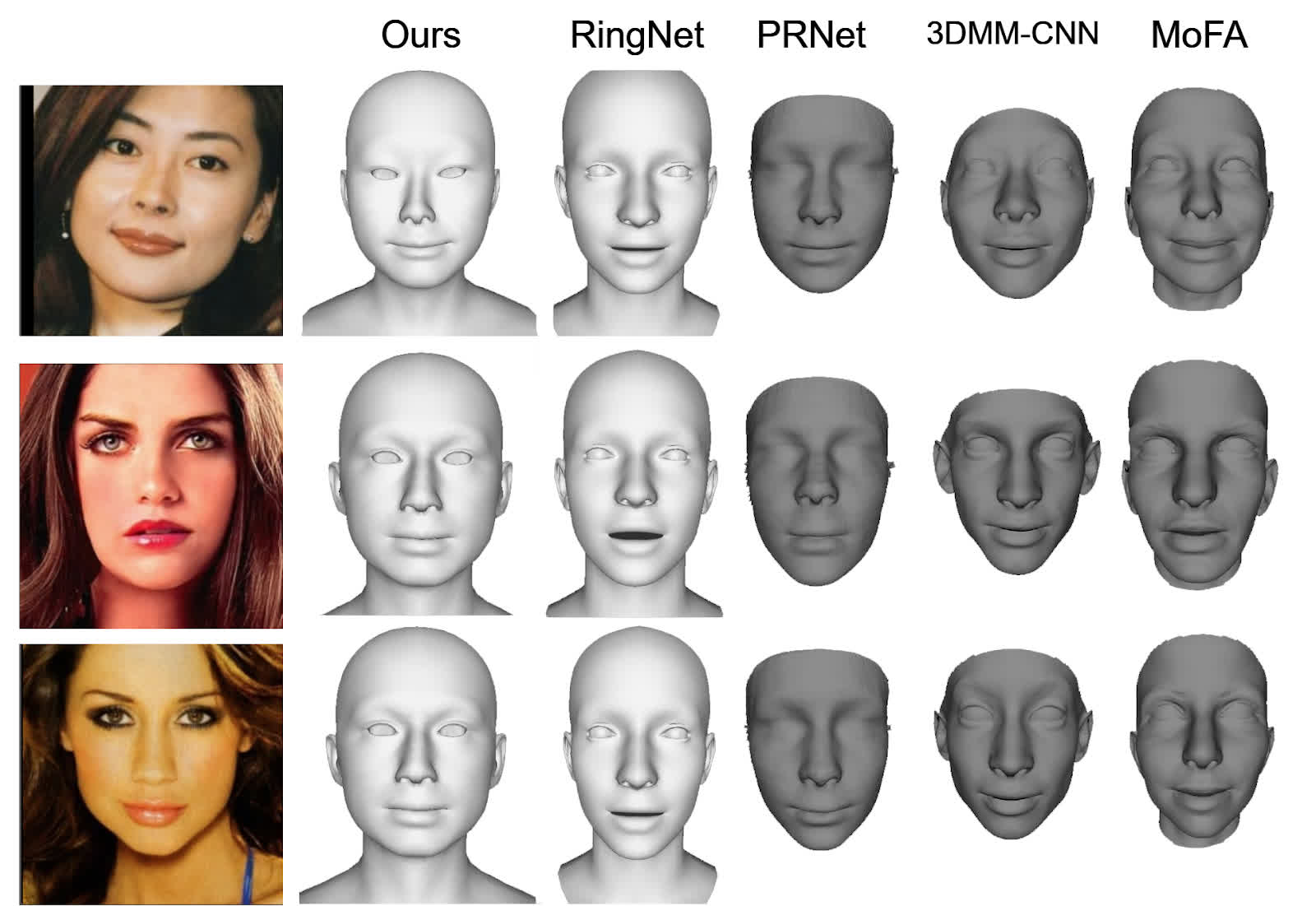}}\\
        \caption{Qualitative results of MoFa dataset. From left to right, input image, our method, RingNet \cite{sanyal2019learning}, PRnet \cite{feng2018prn}, 3DMM-CNN\cite{tuan2017regressing}, and MoFA \cite{Tewari2017MoFAMD}.}
        \label{fig:mofa_qual}
    \end{figure*}

\textbf{More Qualitative Results of MoFA}
    In Fig. \ref{fig:mofa_qual}, we have requested the results from MoFA \cite{Tewari2017MoFAMD} for side-by-side comparisons. Although the quality of reconstructed models are not as good as the results from other database due to the image resolution, large head pose variation, occlusion such as hair and glasses, our model is still considerably better than other methods.

\begin{figure*}
        \centering
        \includegraphics[width= 0.73 \linewidth]{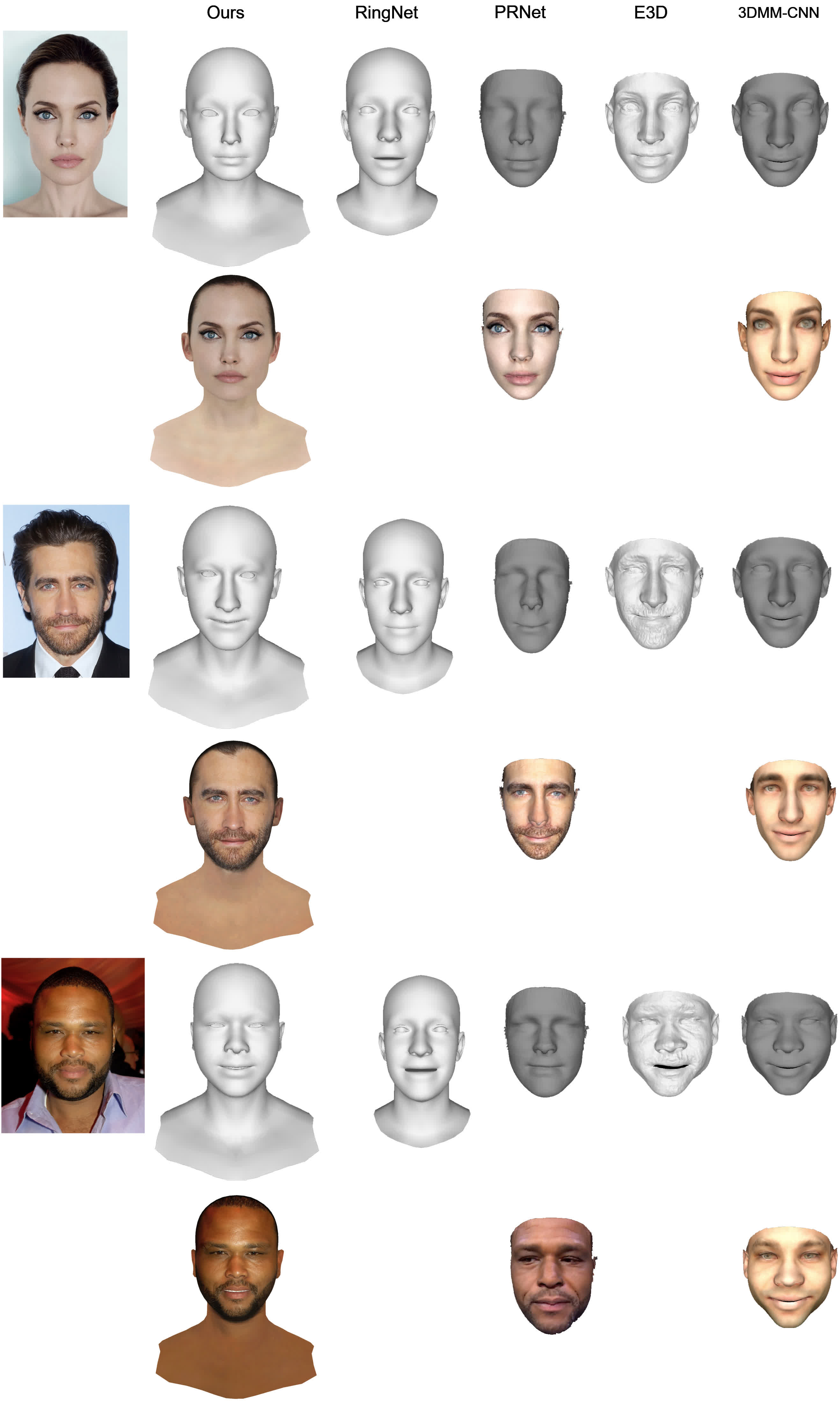}
        \caption{Qualitative results of our method compare to RingNet \cite{sanyal2019learning}, PRnet \cite{feng2018prn}, E3D \cite{tuan2018extreme}, and 3DMM-CNN \cite{tuan2017regressing}. }
        \label{fig:cele_01}
    \end{figure*}

\begin{figure*}
        \centering
        \includegraphics[width= 0.75 \linewidth]{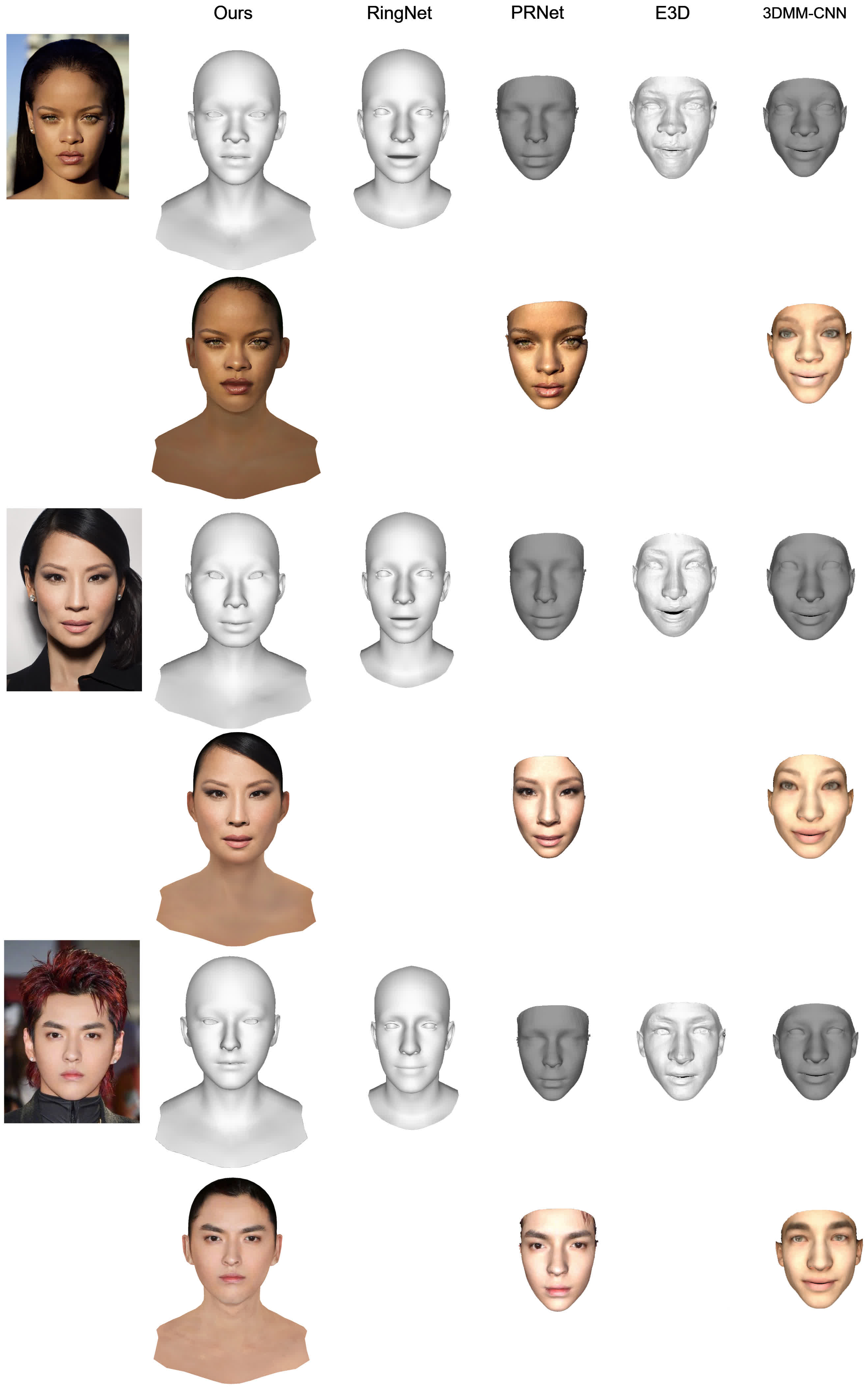}
        \caption{Qualitative results of our method compare to RingNet \cite{sanyal2019learning}, PRnet \cite{feng2018prn}, E3D \cite{tuan2018extreme}, and 3DMM-CNN \cite{tuan2017regressing}.}
        \label{fig:cele_02}
    \end{figure*}

\begin{figure*}
        \centering
        \includegraphics[width= 0.75 \linewidth]{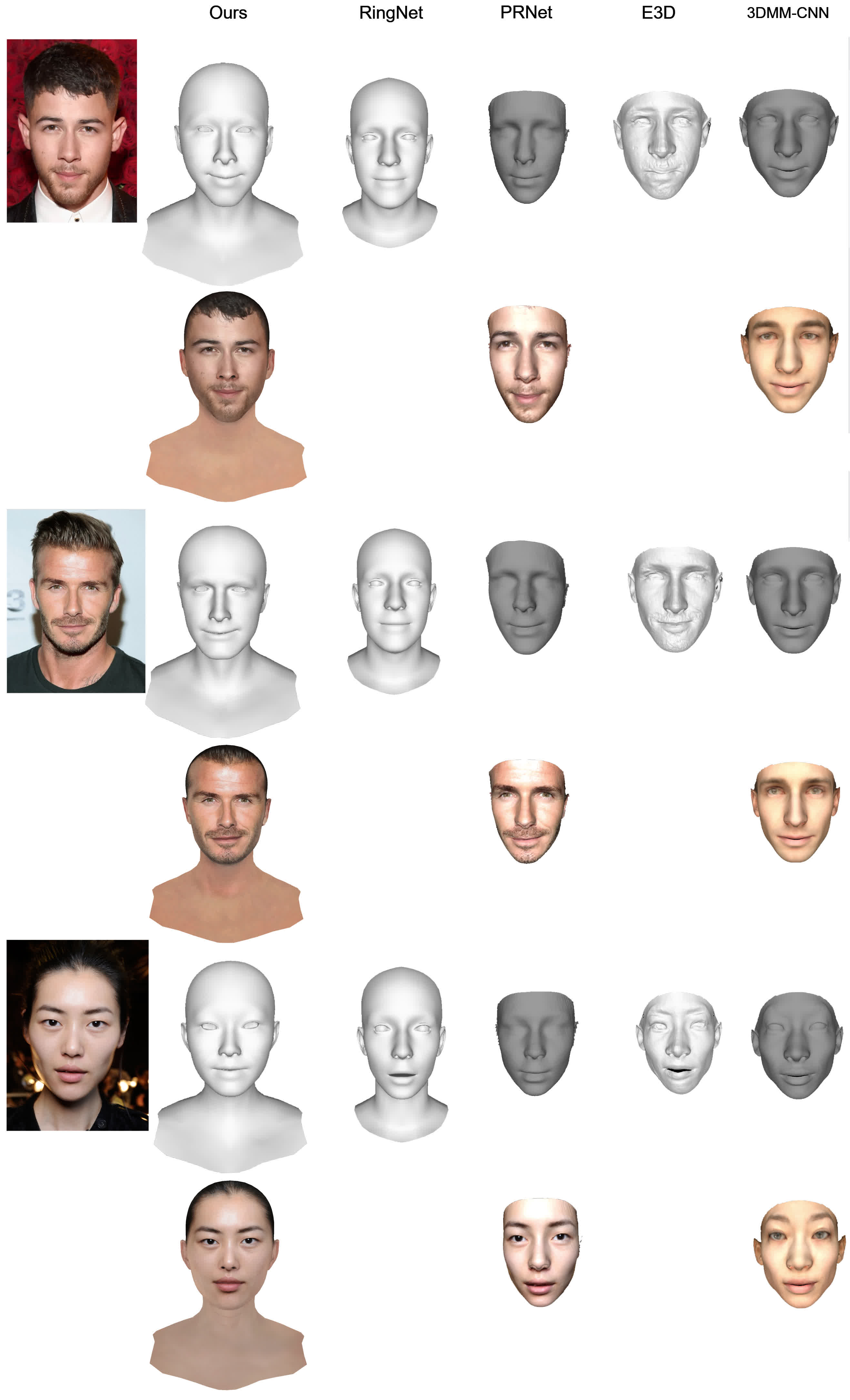}
        \caption{Qualitative results of our method compare to RingNet \cite{sanyal2019learning}, PRnet \cite{feng2018prn}, E3D \cite{tuan2018extreme}, and 3DMM-CNN \cite{tuan2017regressing}.}
        \label{fig:cele_03}
    \end{figure*}

\begin{figure*}
        \centering
        \includegraphics[width= 0.75 \linewidth]{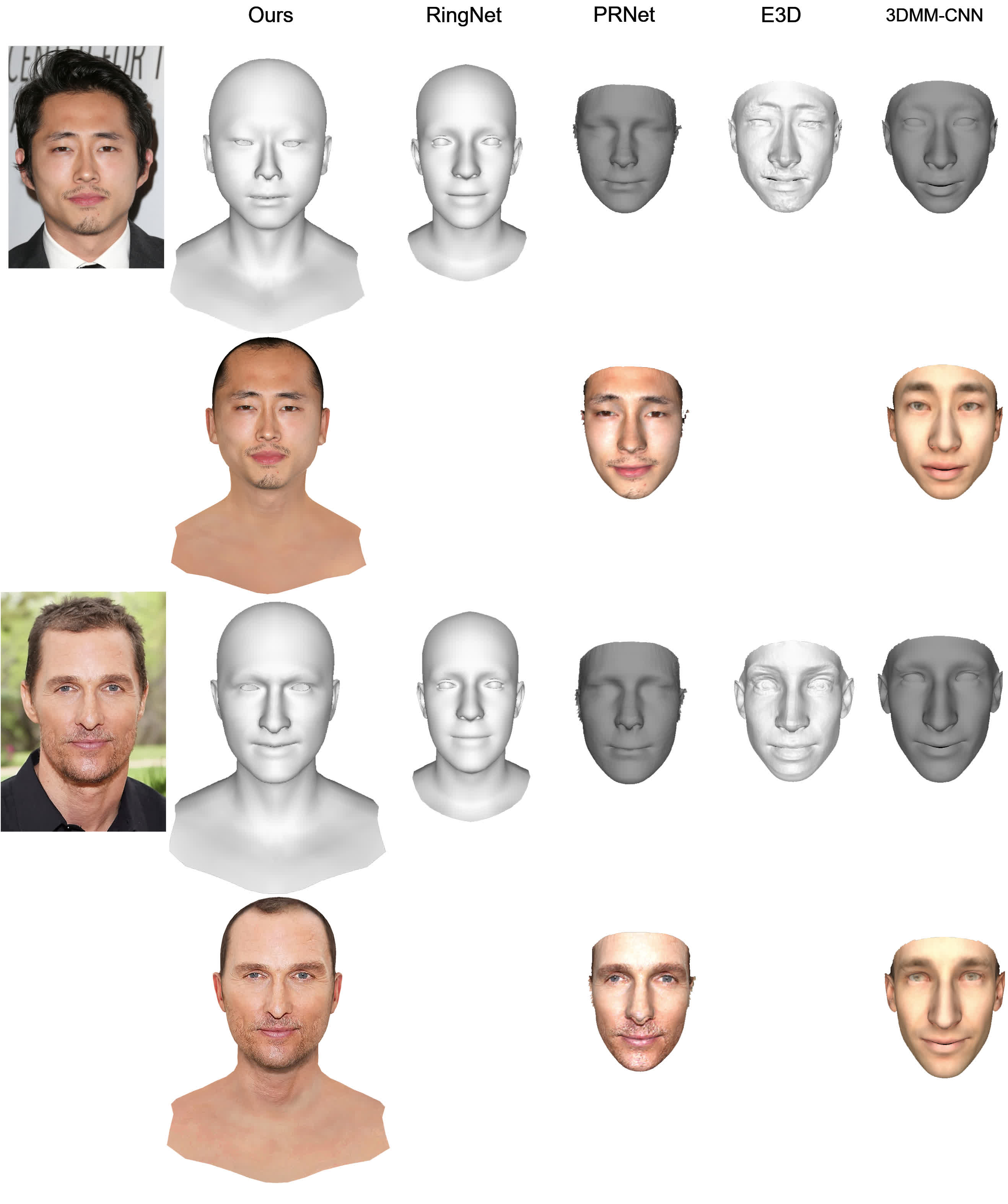}
        \caption{Qualitative results of our method compare to RingNet \cite{sanyal2019learning}, PRnet \cite{feng2018prn}, E3D \cite{tuan2018extreme}, and 3DMM-CNN \cite{tuan2017regressing}.}
        \label{fig:cele_04}
    \end{figure*}

\textbf{More Celebrity-In-the-Wild Results}
    In Figs. \ref{fig:cele_01} - \ref{fig:cele_04}, we present the results of several celebrities and compare our method not only for geometry but also in appearance. Note that by projecting the selfie to a high-resolution UV texture, our reconstructed models has photo-realistic appearance while 3DMM-CNN \cite{tuan2017regressing} and PRNet \cite{feng2018prn} used vertex color results in limited texture reapplication.
    
\section*{Application - Audio-driven Avatar Animation}

\begin{figure*}
        \centering
        \includegraphics[width= 0.6 \linewidth]{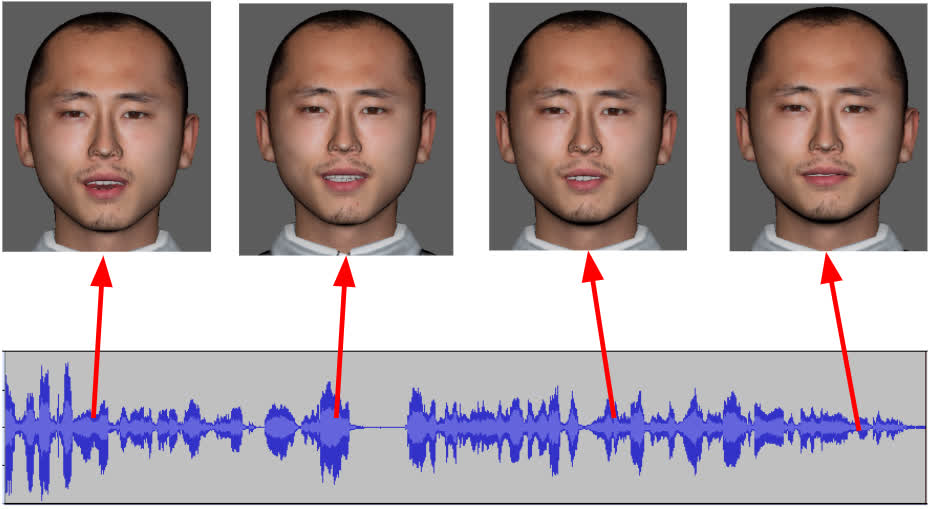}
        \caption{Audio driven lip syncing on our production ready head model}
        \label{fig:audio_driven}
    \end{figure*}

    % In this section, we have some screenshot of using the reconstructed avatar on a mobile device, proven that our model is a ready-to-use, end-to-end selfie to avatar system. More detail of the application can be found in our video.  
    Our automatically generated head model is ready for different applications. Here we demonstrate a case of automatic lip syncing driven by a raw waveform audio input as shown in Fig.~\ref{fig:audio_driven}. For data collection and deep neural network structure, we adopt a similar pipeline as that of \cite{Karras2017} to drive the reconstructed model. All the animation blendshapes are transferred to our generic topology. Please refer to our video for more details. 
    
% \textbf{Face Tracking}
%     We import the created 3D model into Unity and use Faceware \footnote{\url{https://www.facewaretech.com/}} to track the facial movement and create the blendshapes to driven the model animation in real-time. In Fig.~\ref{fig:face_tracking}, the original frame is shown on the left-side  for frame-by-frame comparison.   

%     \begin{figure}
%         \centering
%         \includegraphics[width= 0.78 \linewidth]{Figures/SupplemetalMaterial/audio-driven.png}
%         \caption{The screenshot of face tracking animation.}
%         \label{fig:face_tracking}
%     \end{figure}

%% file: main.bbl
\begin{thebibliography}{10}\itemsep=-1pt

\bibitem{aitpayev2012creation}
Kairat Aitpayev and Jaafar Gaber.
\newblock Creation of 3d human avatar using kinect.
\newblock {\em Asian Transactions on Fundamentals of Electronics, Communication
  \& Multimedia}, 1(5):12--24, 2012.

\bibitem{aldrian2012inverse}
Oswald Aldrian and William~AP Smith.
\newblock Inverse rendering of faces with a 3d morphable model.
\newblock {\em IEEE transactions on pattern analysis and machine intelligence},
  35(5):1080--1093, 2012.

\bibitem{Amberg2007_nICP}
Brian Amberg, Sami Romdhani, and Thomas Vetter.
\newblock Optimal step nonrigid icp algorithms for surface registration.
\newblock In {\em Computer Vision and Pattern Recognition, 2007. CVPR'07. IEEE
  Conference on}, pages 1--8. IEEE, 2007.

\bibitem{Blanz1999_3DMD}
Volker Blanz and Thomas Vetter.
\newblock A morphable model for the synthesis of 3d faces.
\newblock In {\em Proceedings of the 26th Annual Conference on Computer
  Graphics and Interactive Techniques}, SIGGRAPH '99, pages 187--194, New York,
  NY, USA, 1999. ACM Press/Addison-Wesley Publishing Co.

\bibitem{Bolkart2015_GML}
T. {Bolkart} and S. {Wuhrer}.
\newblock A groupwise multilinear correspondence optimization for 3d faces.
\newblock In {\em 2015 IEEE International Conference on Computer Vision
  (ICCV)}, pages 3604--3612, Dec 2015.

\bibitem{Booth2016_3DMM10000}
J. {Booth}, A. {Roussos}, S. {Zafeiriou}, A. {Ponniahy}, and D. {Dunaway}.
\newblock A 3d morphable model learnt from 10,000 faces.
\newblock In {\em 2016 IEEE Conference on Computer Vision and Pattern
  Recognition (CVPR)}, pages 5543--5552, June 2016.

\bibitem{cao2014displaced}
Chen Cao, Qiming Hou, and Kun Zhou.
\newblock Displaced dynamic expression regression for real-time facial tracking
  and animation.
\newblock {\em ACM Transactions on graphics (TOG)}, 33(4):43, 2014.

\bibitem{Cao2014_FWH}
C. {Cao}, Y. {Weng}, S. {Zhou}, Y. {Tong}, and K. {Zhou}.
\newblock Facewarehouse: A 3d facial expression database for visual computing.
\newblock {\em IEEE Transactions on Visualization and Computer Graphics},
  20(3):413--425, March 2014.

\bibitem{debevec2000acquiring}
Paul Debevec, Tim Hawkins, Chris Tchou, Haarm-Pieter Duiker, Westley Sarokin,
  and Mark Sagar.
\newblock Acquiring the reflectance field of a human face.
\newblock In {\em Proceedings of the 27th annual conference on Computer
  graphics and interactive techniques}, pages 145--156. ACM
  Press/Addison-Wesley Publishing Co., 2000.

\bibitem{dou2017end}
Pengfei Dou, Shishir~K Shah, and Ioannis~A Kakadiaris.
\newblock End-to-end 3d face reconstruction with deep neural networks.
\newblock In {\em Proceedings of the IEEE Conference on Computer Vision and
  Pattern Recognition}, pages 5908--5917, 2017.

\bibitem{feng2018prn}
Yao Feng, Fan Wu, Xiaohu Shao, Yanfeng Wang, and Xi Zhou.
\newblock Joint 3d face reconstruction and dense alignment with position map
  regression network.
\newblock In {\em ECCV}, 2018.

\bibitem{Feng2018esrc}
Z. {Feng}, P. {Huber}, J. {Kittler}, P. {Hancock}, X. {Wu}, Q. {Zhao}, P.
  {Koppen}, and M. {Raetsch}.
\newblock Evaluation of dense 3d reconstruction from 2d face images in the
  wild.
\newblock In {\em 2018 13th IEEE International Conference on Automatic Face
  Gesture Recognition (FG 2018)}, pages 780--786, May 2018.

\bibitem{gao2019sparse}
Lin Gao, Yu-Kun Lai, Jie Yang, Zhang Ling-Xiao, Shihong Xia, and Leif Kobbelt.
\newblock Sparse data driven mesh deformation.
\newblock {\em IEEE transactions on visualization and computer graphics}, 2019.

\bibitem{gecer2019ganfit}
Baris Gecer, Stylianos Ploumpis, Irene Kotsia, and Stefanos Zafeiriou.
\newblock Ganfit: Generative adversarial network fitting for high fidelity 3d
  face reconstruction.
\newblock In {\em Proceedings of the IEEE Conference on Computer Vision and
  Pattern Recognition}, pages 1155--1164, 2019.

\bibitem{geng20193d}
Zhenglin Geng, Chen Cao, and Sergey Tulyakov.
\newblock 3d guided fine-grained face manipulation.
\newblock In {\em Proceedings of the IEEE Conference on Computer Vision and
  Pattern Recognition}, pages 9821--9830, 2019.

\bibitem{genova2018unsupervised}
Kyle Genova, Forrester Cole, Aaron Maschinot, Aaron Sarna, Daniel Vlasic, and
  William~T Freeman.
\newblock Unsupervised training for 3d morphable model regression.
\newblock In {\em Proceedings of the IEEE Conference on Computer Vision and
  Pattern Recognition}, pages 8377--8386, 2018.

\bibitem{hu2017avatar}
Liwen Hu, Shunsuke Saito, Lingyu Wei, Koki Nagano, Jaewoo Seo, Jens Fursund,
  Iman Sadeghi, Carrie Sun, Yen-Chun Chen, and Hao Li.
\newblock Avatar digitization from a single image for real-time rendering.
\newblock {\em ACM Transactions on Graphics (TOG)}, 36(6):195, 2017.

\bibitem{li2018avatar}
Liwen Hu, Shunsuke Saito, Lingyu Wei, Koki Nagano, Jaewoo Seo, Jens Fursund,
  Iman Sadeghi, Carrie Sun, Yen-Chun Chen, and Hao Li.
\newblock Avatar digitization from a single image for real-time rendering, nov
  2017.

\bibitem{ichim2015dynamic}
Alexandru~Eugen Ichim, Sofien Bouaziz, and Mark Pauly.
\newblock Dynamic 3d avatar creation from hand-held video input.
\newblock {\em ACM Transactions on Graphics (ToG)}, 34(4):45, 2015.

\bibitem{jackson2017large}
Aaron~S Jackson, Adrian Bulat, Vasileios Argyriou, and Georgios Tzimiropoulos.
\newblock Large pose 3d face reconstruction from a single image via direct
  volumetric cnn regression.
\newblock In {\em Proceedings of the IEEE International Conference on Computer
  Vision}, pages 1031--1039, 2017.

\bibitem{jiang2019disentangled}
Zi-Hang Jiang, Qianyi Wu, Keyu Chen, and Juyong Zhang.
\newblock Disentangled representation learning for 3d face shape.
\newblock {\em Proceedings of the IEEE Conference on Computer Vision and
  Pattern Recognition}, 2018.

\bibitem{jourabloo2016large}
Amin Jourabloo and Xiaoming Liu.
\newblock Large-pose face alignment via cnn-based dense 3d model fitting.
\newblock In {\em Proceedings of the IEEE conference on computer vision and
  pattern recognition}, pages 4188--4196, 2016.

\bibitem{Karras2017}
Tero Karras, Timo Aila, Samuli Laine, Antti Herva, and Jaakko Lehtinen.
\newblock Audio-driven facial animation by joint end-to-end learning of pose
  and emotion.
\newblock {\em ACM Trans. Graph.}, 36(4):94:1--94:12, July 2017.

\bibitem{kazemi2014one}
Vahid Kazemi and Josephine Sullivan.
\newblock One millisecond face alignment with an ensemble of regression trees.
\newblock In {\em Proceedings of the IEEE conference on computer vision and
  pattern recognition}, pages 1867--1874, 2014.

\bibitem{Koppen2018GMM}
Paul Koppen, Zhen-Hua Feng, Josef Kittler, Muhammad Awais, William Christmas,
  Xiao-Jun Wu, and He-Feng Yin.
\newblock Gaussian mixture 3d morphable face model.
\newblock {\em Pattern Recogn.}, 74(C):617--628, Feb 2018.

\bibitem{lepetit2009epnp}
Vincent Lepetit, Francesc Moreno-Noguer, and Pascal Fua.
\newblock Epnp: An accurate o (n) solution to the pnp problem.
\newblock {\em International journal of computer vision}, 81(2):155, 2009.

\bibitem{Li2017_4D}
Tianye Li, Timo Bolkart, Michael.~J. Black, Hao Li, and Javier Romero.
\newblock Learning a model of facial shape and expression from {4D} scans.
\newblock {\em ACM Transactions on Graphics, (Proc. SIGGRAPH Asia)}, 36(6),
  2017.

\bibitem{parkhi2015deep}
Omkar~M. Parkhi, Andrea Vedaldi, and Andrew Zisserman.
\newblock Deep face recognition.
\newblock In Mark W.~Jones Xianghua~Xie and Gary K.~L. Tam, editors, {\em
  Proceedings of the British Machine Vision Conference (BMVC)}, pages
  41.1--41.12. BMVA Press, September 2015.

\bibitem{Paysan2009_BFM}
P. {Paysan}, R. {Knothe}, B. {Amberg}, S. {Romdhani}, and T. {Vetter}.
\newblock A 3d face model for pose and illumination invariant face recognition.
\newblock In {\em 2009 Sixth IEEE International Conference on Advanced Video
  and Signal Based Surveillance}, pages 296--301, Sep. 2009.

\bibitem{perez2003poisson}
Patrick P{\'e}rez, Michel Gangnet, and Andrew Blake.
\newblock Poisson image editing.
\newblock {\em ACM Transactions on graphics (TOG)}, 22(3):313--318, 2003.

\bibitem{romdhani2005estimating}
Sami Romdhani and Thomas Vetter.
\newblock Estimating 3d shape and texture using pixel intensity, edges,
  specular highlights, texture constraints and a prior.
\newblock In {\em 2005 IEEE Computer Society Conference on Computer Vision and
  Pattern Recognition (CVPR'05)}, volume~2, pages 986--993. IEEE, 2005.

\bibitem{sanyal2019learning}
Soubhik Sanyal, Timo Bolkart, Haiwen Feng, and Michael~J. Black.
\newblock Learning to regress 3d face shape and expression from an image
  without 3d supervision, 2019.

\bibitem{schroff2015facenet}
Florian Schroff, Dmitry Kalenichenko, and James Philbin.
\newblock Facenet: A unified embedding for face recognition and clustering.
\newblock In {\em Proceedings of the IEEE conference on computer vision and
  pattern recognition}, pages 815--823, 2015.

\bibitem{sela2017unrestricted}
Matan Sela, Elad Richardson, and Ron Kimmel.
\newblock Unrestricted facial geometry reconstruction using image-to-image
  translation.
\newblock {\em arxiv}, 2017.

\bibitem{sengupta2018sfsnet}
Soumyadip Sengupta, Angjoo Kanazawa, Carlos~D Castillo, and David~W Jacobs.
\newblock Sfsnet: Learning shape, reflectance and illuminance of facesin the
  wild'.
\newblock In {\em Proceedings of the IEEE Conference on Computer Vision and
  Pattern Recognition}, pages 6296--6305, 2018.

\bibitem{sorkine2004laplacian}
Olga Sorkine, Daniel Cohen-Or, Yaron Lipman, Marc Alexa, Christian R{\"o}ssl,
  and H-P Seidel.
\newblock Laplacian surface editing.
\newblock In {\em Proceedings of the 2004 Eurographics/ACM SIGGRAPH symposium
  on Geometry processing}, pages 175--184. ACM, 2004.

\bibitem{taigman2014deepface}
Yaniv Taigman, Ming Yang, Marc'Aurelio Ranzato, and Lior Wolf.
\newblock Deepface: Closing the gap to human-level performance in face
  verification.
\newblock In {\em Proceedings of the IEEE conference on computer vision and
  pattern recognition}, pages 1701--1708, 2014.

\bibitem{tewari2018self}
Ayush Tewari, Michael Zollh{\"o}fer, Pablo Garrido, Florian Bernard, Hyeongwoo
  Kim, Patrick P{\'e}rez, and Christian Theobalt.
\newblock Self-supervised multi-level face model learning for monocular
  reconstruction at over 250 hz.
\newblock In {\em Proceedings of the IEEE Conference on Computer Vision and
  Pattern Recognition}, pages 2549--2559, 2018.

\bibitem{Tewari2017MoFAMD}
Ayush Tewari, Michael Zollh{\"o}fer, Hyeongwoo Kim, Pablo Garrido, Florian
  Bernard, Patrick P{\'e}rez, and Christian Theobalt.
\newblock Mofa: Model-based deep convolutional face autoencoder for
  unsupervised monocular reconstruction.
\newblock {\em 2017 IEEE International Conference on Computer Vision (ICCV)},
  pages 3735--3744, 2017.

\bibitem{thies2016face2face}
Justus Thies, Michael Zollhofer, Marc Stamminger, Christian Theobalt, and
  Matthias Nie{\ss}ner.
\newblock Face2face: Real-time face capture and reenactment of rgb videos.
\newblock In {\em Proceedings of the IEEE Conference on Computer Vision and
  Pattern Recognition}, pages 2387--2395, 2016.

\bibitem{Tran2018_nl3DMM}
Luan Tran and Xiaoming Liu.
\newblock Nonlinear 3d face morphable model.
\newblock In {\em IEEE Computer Vision and Pattern Recognition (CVPR)}, Salt
  Lake City, UT, June 2018.

\bibitem{tuan2017regressing}
Anh Tuan~Tran, Tal Hassner, Iacopo Masi, and G{\'e}rard Medioni.
\newblock Regressing robust and discriminative 3d morphable models with a very
  deep neural network.
\newblock In {\em Proceedings of the IEEE Conference on Computer Vision and
  Pattern Recognition}, pages 5163--5172, 2017.

\bibitem{tuan2018extreme}
Anh Tuan~Tran, Tal Hassner, Iacopo Masi, Eran Paz, Yuval Nirkin, and G{\'e}rard
  Medioni.
\newblock Extreme 3d face reconstruction: Seeing through occlusions.
\newblock In {\em Proceedings of the IEEE Conference on Computer Vision and
  Pattern Recognition}, pages 3935--3944, 2018.

\bibitem{Vlasic2005_FTM}
Daniel Vlasic, Matthew Brand, Hanspeter Pfister, and Jovan Popovi\'{c}.
\newblock Face transfer with multilinear models.
\newblock {\em ACM Trans. Graph.}, 24(3):426--433, jul 2005.

\bibitem{wang2011text}
Lijuan Wang, Wei Han, Frank~K Soong, and Qiang Huo.
\newblock Text driven 3d photo-realistic talking head.
\newblock In {\em Twelfth Annual Conference of the International Speech
  Communication Association}, 2011.

\bibitem{wu2018alive}
Qianyi Wu, Juyong Zhang, Yu-Kun Lai, Jianmin Zheng, and Jianfei Cai.
\newblock Alive caricature from 2d to 3d.
\newblock In {\em Proceedings of the IEEE Conference on Computer Vision and
  Pattern Recognition}, pages 7336--7345, 2018.

\bibitem{wu2018light}
Xiang Wu, Ran He, Zhenan Sun, and Tieniu Tan.
\newblock A light cnn for deep face representation with noisy labels.
\newblock {\em IEEE Transactions on Information Forensics and Security},
  13(11):2884--2896, 2018.

\bibitem{yamaguchi2018high}
Shuco Yamaguchi, Shunsuke Saito, Koki Nagano, Yajie Zhao, Weikai Chen, Kyle
  Olszewski, Shigeo Morishima, and Hao Li.
\newblock High-fidelity facial reflectance and geometry inference from an
  unconstrained image.
\newblock {\em ACM Transactions on Graphics (TOG)}, 37(4):162, 2018.

\bibitem{yi2019mmface}
Hongwei Yi, Chen Li, Qiong Cao, Xiaoyong Shen, Sheng Li, Guoping Wang, and
  Yu-Wing Tai.
\newblock Mmface: A multi-metric regression network for unconstrained face
  reconstruction.
\newblock In {\em Proceedings of the IEEE Conference on Computer Vision and
  Pattern Recognition}, pages 7663--7672, 2019.

\bibitem{zhu2016face}
Xiangyu Zhu, Zhen Lei, Xiaoming Liu, Hailin Shi, and Stan~Z Li.
\newblock Face alignment across large poses: A 3d solution.
\newblock In {\em Proceedings of the IEEE conference on computer vision and
  pattern recognition}, pages 146--155, 2016.

\end{thebibliography}
